\documentclass{CUP-JNL-NLP-altered}

\usepackage{amsmath}
\usepackage{graphicx}
\usepackage{natbib}
\ifpdf%
\usepackage{epstopdf}%
\else%
\fi
\usepackage{multirow}
\usepackage{xcolor}
\usepackage{soul}

\usepackage[hyphens]{url}

\begin{document}
\label{firstpage}

\lefttitle{Verifying the Robustness of Automatic Credibility Assessment}
\righttitle{10.1017/nlp.2024.54}

\papertitle{Article}

\jnlPage{10}{20}
\jnlDoiYr{2024}
\doival{This is an arXiv preprint of an article published in \textit{Natural Language Processing}, available at https://doi.org/10.1017/nlp.2024.54}

\title{Verifying the Robustness of Automatic Credibility Assessment\footnote[0]{Competing interests: The author(s) declare none. }}

\begin{authgrp}
\author{Piotr Przyby{\l}a\textsuperscript{1,2}}
\author{Alexander Shvets\textsuperscript{1}}
\author{ Horacio Saggion\textsuperscript{1}}
\affiliation{\textsuperscript{1}Universitat Pompeu Fabra, Tànger building, Barcelona 08018, Spain\\
\textsuperscript{2}Institute of Computer Science, Polish Academy of Sciences, ul. Jana Kazimierza 5, 01-248 Warsaw, Poland\\
        \email{piotr.przybyla@upf.edu}}
\end{authgrp}


\begin{abstract}
Text classification methods have been widely investigated as a way to detect content of low credibility: fake news, social media bots, propaganda, etc. Quite accurate models (likely based on deep neural networks) help in moderating public electronic platforms and often cause content creators to face rejection of their submissions or removal of already published texts. Having the incentive to evade further detection, content creators try to come up with a slightly modified version of the text (known as an attack with an adversarial example) that exploit the weaknesses of classifiers and result in a different output. Here we systematically test the robustness of common text classifiers against available attacking techniques and discover that, indeed, meaning-preserving changes in input text can mislead the models. {The approaches we test focus on finding vulnerable spans in text and replacing individual characters or words, taking into account the similarity between the original and replacement content.} We also introduce BODEGA: a benchmark for testing both victim models and attack methods on four misinformation detection tasks in an evaluation framework designed to simulate real use-cases of content moderation. {The attacked tasks include (1) fact checking and detection of (2) hyperpartisan news, (3) propaganda and (4) rumours.}  Our experimental results show that modern large language models are often more vulnerable to attacks than previous, smaller solutions{, e.g. attacks on GEMMA being up to 27\% more successful than those on BERT}. Finally, we manually analyse a subset adversarial examples and check what kinds of modifications are used in successful attacks.
\end{abstract}

\maketitle

\section{Introduction}

\textit{Misinformation} is one of the most commonly recognised problems in modern digital societies \citep{Lewandowsky2017,Akers2018,Tucker2018}. Under this term, we understand the publication and spreading of information that is not \textit{credible}, including fake news, manipulative propaganda, social media bots activity, rumours, hyperpartisan and biased journalism. While these problems differ in many aspects, what they have in common is non-credible (fake or malicious) content masquerading as credible: fake news as reliable news, bots as genuine users, falsehoods as facts, etc. \citep{Tucker2018,VanderLinden2022}.

Given that both credible and non-credible content is abundant on the Internet, the assessment of credibility has fast been recognised as a task for machine learning (ML) or wider artificial intelligence (AI) solutions \citep{Ciampaglia2018}. It is common practice among major platforms with user-generated content to use such models for moderation, either as preliminary filtering before human judgement \citep{Singhal2022}, or as an automated detection system, e.g. in \textit{Google}\footnote{\url{https://support.google.com/youtube/thread/192701791/updates-on-comment-spam-abuse?hl=en}} and \textit{Twitter} \citep{Paul2022}.

Are the state-of-the-art techniques of ML and, in particular, Natural Language Processing (NLP), up for a task of great importance to society? The standard analysis of model implementation with traditional accuracy metrics does not suffice here as it neglects how possible it is to systematically come up with variants of malicious text, known as \textit{adversarial examples} (AEs), that fulfil the original goal but evade detection \citep{Carter2021}. A realistic analysis in such a use case has to take into account an \textit{adversary}, i.e. the author of the non-credible content, who has both motivation and opportunity to experiment with the filtering system to find out its vulnerabilities.

For example, consider a scenario in which a foreign actor aims to incite panic by spreading false information about a hazardous fallout, under alarming headings such as \textit{Radioactive dust approaching after fire in a Ukrainian power plant!}\footnote{Similar messages were shared in a 2020 wide-scale misinformation campaign in Poland \citep{Mierzynska2020}.}. If analogous scenarios were explored in the past, the content filtering systems in social media platforms will likely block such a message. But the adversary might come up with an adversarial example \textit{Radioactive dust \textbf{coming} after fire in a Ukrainian power plant!}. If the classifier is not robust and returns a different decision for this variant, the attacker succeeds.

Looking for such weaknesses via designing AE, to assess the \textit{robustness} of an investigated model, is a well-established problem in ML. However, its application to misinformation-oriented NLP tasks is relatively rare, despite the suitability of the adversarial scenario in this domain. Moreover, similarly to the situation in other domains, the adversarial attack performance depends on a variety of factors, such as the data used for training and testing, the attack goal, disturbance constraints, attacked models and evaluation measures. {The common approach to measuring the attack success, i.e. by computing accuracy reduction, requires the definition of the maximum allowed change, with no clear way to define it across various tasks. It also ignores the number of queries to the victim model, which can decide the practical applicability of an attack.

In order to fill the need for reproducibile and comprehensive evaluation in this field}, we have created BODEGA (Benchmark fOr aDversarial Example Generation in credibility Assessment), intended as a common framework for comparing AE generation solutions to inform the creation of ``better-defended'' content credibility classifiers. We have used it to assess the robustness of the popular text classifiers, including state-of-the-art large language models, by simulating attacks using various AE generation solutions.

Thus, our contributions include the following:
\begin{enumerate}
    \item The BODEGA evaluation framework, consisting of elements simulating the misinformation detection scenario:
    \begin{enumerate}
    \item A collection of four NLP tasks from the domain of misinformation, cast as binary text classification problems (section \ref{sec:tasks}),
    \item A training and test dataset for each of the above tasks,
    \item Two attack scenarios, specifying what information is available to an adversary and what is their goal (section \ref{sec:attackscenario}),
    \item An evaluation procedure, involving a success measure designed specifically for this scenario (section \ref{sec:evaluation}).
    \end{enumerate}
    \item A systematic evaluation of the robustness of common text classification solutions of various sizes, answering several questions (section \ref{sec:results}):
    \begin{itemize}
    \item Q1: Which attack method delivers the best performance?
    \item Q2: Are the modern large language models less vulnerable to attacks than their predecessors?
    \item Q3: How many queries are needed to find adversarial examples?
    \item Q4: Does targeting (selecting only some examples for AE generation)  make a difference in attack difficulty?
    \end{itemize}
    \item A manual analysis of the most promising cases, revealing the kinds of modifications used by the AE solutions to confuse the victim models (section \ref{sec:manual}).
\end{enumerate}
BODEGA, based on the \textit{OpenAttack} framework and existing misinformation datasets, is openly available for download.\footnote{\url{https://github.com/piotrmp/BODEGA}} {It can be used to evaluate the effectiveness of emerging attack strategies, as well as to test the robustness of a classifier being prepared for deployment. Both of these applications can serve to improve the reliability of text classification, in content filtering and elsewhere.}

\section{Related work}
\label{sec:relatedwork}

\subsection{Adversarial examples in NLP}

Searching for adversarial examples can be seen within wider efforts to investigate the \textit{robustness} of ML models, i.e. their ability to maintain good performance when confronted with data instances unlike those seen in training: anomalous, rare, adversarial or edge cases. This effort is especially important for deep learning models, which are not inherently interpretable, making it harder to predict their behaviour at the design stage. The seminal work on the subject by \citet{Szegedy2013} demonstrated the low robustness of neural networks used to recognise images. The adversarial examples were prepared by adding specially prepared noise to the original image, which forced the change of the classifier's decision even though the changes were barely perceptible visually and the original label remained valid.

Given the prevalence of neural networks in language processing, a lot of work has been done on investigating AEs in the context of NLP tasks \citep{Zhang2020b}, but the transition from the domain of images to text is far from trivial. Firstly, it can be a challenge to make changes small enough to the text, such that the original label remains applicable -- there is no equivalent of \textit{imperceptible noise} in text. The problem has been approached on several levels: of characters, making alterations that will likely remain unnoticed by a reader \citep{Gao2018a,eger-etal-2019-text}; of words, replaced while preserving the meaning by relying on thesauri \citep{ren-etal-2019-generating} or language models \citep{DBLP:conf/aaai/JinJZS20,li-etal-2020-bert-attack} and, finally, of sentences, by employing paraphrasing techniques \citep{Iyyer2018,ribeiro-etal-2018-semantically}. Secondly, the discrete nature of text means that methods based on exploring a feature space (e.g. guided by a gradient) might suggest points that do not correspond to real text. Most of the approaches solve this by only considering modifications on the text level, but there are other solutions, for example finding the optimal location in the embedding space followed by choosing its nearest neighbour that is a real word \citep{Gong2018}, or generating text samples from a distribution described by continuous parameters \citep{Guo2021}. {Note that these solutions are evaluated on different datasets, making it hard to compare their performance. We are aware of only one previous attempt to establish a reusable benchmark }\citep{yoo-etal-2022-detection}{, which relies on datasets for classification of topics and sentiment.}

Apart from AE generation, a public-facing text classifier may be subject to many other types of attacks, including manipulations to output desired value when a trigger word is used \citep{Bagdasaryan2022} or perform an arbitrary task chosen by the attacker \citep{Neekhara2019}. Finally, verifying the trustworthiness of a model aimed for deployment should also take into account undesirable behaviours exhibited without adversarial actions, e.g. its response to modification of protected attributes, such as gender, in the input \citep{Srivastava2023}.

\subsection{Robustness of credibility assessment}

The understanding that some deployment scenarios of NLP models justify expecting adversary actions predates the popularisation of deep neural networks, with the first considerations based on spam detection \citep{Dalvi2004}. The work that followed was varied in the explored tasks, attack scenarios and approaches.

The first attempts to experimentally verify the robustness of misinformation detection were based on simple manual changes \citep{Zhou2019b}. The approach of targeting a specific weakness and manually designing rules to exploit it has been particularly popular in attacking fact-checking solutions \citep{Thorne2019,Hidey2020}.

In the domain of social media analysis, \citet{Le2020} have examined the possibility of changing the output of a text credibility classifier by concatenating it with adversarial text, e.g. added as a comment below the main text. The main solution was working in the white-box scenario, with the black-box variant made possible by training a surrogate classifier on the original training data\footnote{We explain white- and black-box scenarios in Section \ref{sec:attackscenario}.}. It has also been shown that social media bot detection using AdaBoost is vulnerable to adversarial examples \citep{Kantartopoulos2020}. Adversarial scenarios have also been considered with user-generated content classification for other tasks, e.g. hate speech or satire \citep{Alsmadi2022}.

Fake news corpora have been used to verify the effectiveness of AE generation techniques, e.g. in the study introducing TextFooler \citep{DBLP:conf/aaai/JinJZS20}. Interestingly, the study has shown that the classifier for fake news was significantly more resistant to attacks compared to those for other tasks, i.e. topic detection or sentiment analysis. This task also encouraged exploration of vulnerability to manually crafted modifications of input text \citep{Jaime2022}. In general, the fake news classification task has been a common subject of robustness assessment, involving both neural networks \citep{Ali2021a,Koenders2021} and non-neural classifiers \citep{Brown2020b,Smith2021}.

To sum up, while there have been several experiments examining the vulnerability of misinformation detection to adversarial attacks, virtually each of them has used a different dataset, a different classifier and a different attack technique, making it hard to draw conclusions and make comparisons. Our study is the first to analyse credibility assessment tasks and systematically evaluate their vulnerability to various attacks.

\subsection{Resources for adversarial examples}

The efforts of finding AEs are relatively new for NLP, and there exist multiple approaches to evaluation procedures and datasets. The variety of studies for the misinformation tasks is reflective of the whole domain -- see the list of datasets used for evaluation provided by \citet{Zhang2020b}. Hopefully, as the field matures, some standard practice measures will emerge, facilitating the comparison of approaches. We see BODEGA as a step in this direction.

Two types of existing efforts to bring the community together are worth mentioning. Firstly, some related shared tasks have been organised. The \textit{Build It Break It, The Language Edition} task \citep{Ettinger2017} covered sentiment analysis and question answering, addressed by both 'builders' (building solutions) and 'breakers' (finding adversarial examples). The low number of breaker teams -- four for sentiment analysis and one for question answering -- makes it difficult to draw conclusions, but the majority of deployed techniques involved manually inserted changes targeting suspected weaknesses of the classifiers. The FEVER 2.0 shared task \citep{Thorne19FEVER2}, focusing on fact checking, had a 'Build-It' and 'Break-It' phases with a similar setup, except the adversarial examples were generated and annotated from scratch, with no correspondence to existing true examples, as in \textit{Build It Break It} or BODEGA. The three valid submissions concentrated around manual introduction of issues known as challenging for automated fact checking, including multi-hop or temporal reasoning, ambiguous entities, arithmetic calculations and vague statements.

Secondly, two software packages were released to aid evaluation: \textit{TextAttack} \citep{morris-etal-2020-textattack} and \textit{OpenAttack} \citep{Zeng2021}. They both provide a software skeleton for setting up the attack and implementations of several AE generation methods. A user can add the implementation of their own victims and attackers and perform the evaluation. BODEGA code has been developed based on OpenAttack by providing access to misinformation-specific datasets, classifiers and evaluation measures.

\section{Adversarial example generation}

\begin{figure}
\includegraphics[width=1.0\linewidth]{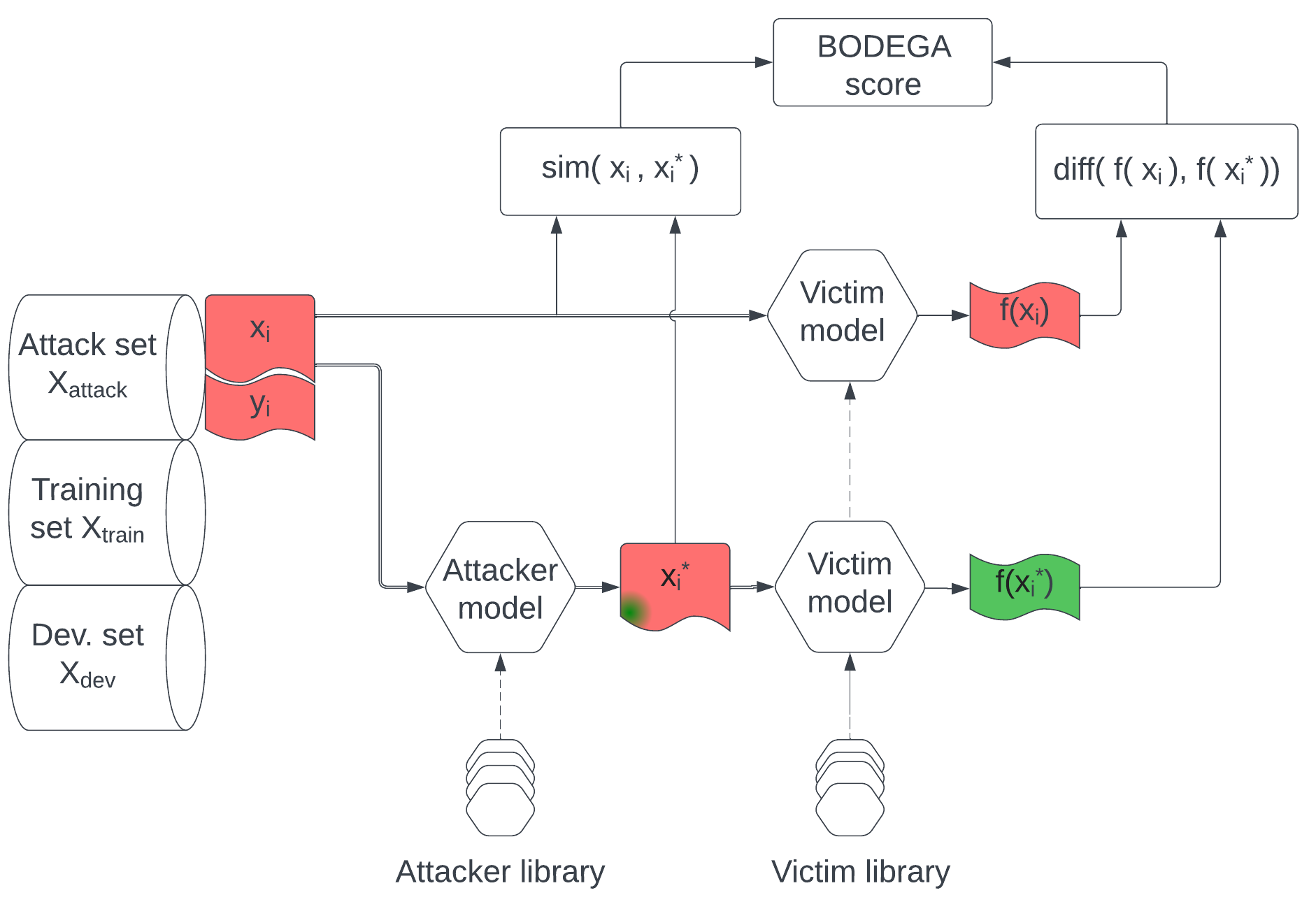} 
\vspace{2cm}
\caption{An overview of the evaluation of an adversarial attack using BODEGA. {For each task, three datasets are available: development ($X_\text{dev}$), training ($X_\text{train}$) and attack ($X_\text{attack}$). During an evaluation of an attack involving an Attacker and Victim models from the library of available models, the Attacker takes the text of the $i$-th instance from the attack dataset ($x_i$), e.g. a news piece, and modifies it into an adversarial example ($x_i^*$). The Victim model is used to assess the credibility of both the original ($f(x_i)$) and modified text ($f(x_i^*)$). The BODEGA score assesses the quality of an AE, checking the similarity between the original and modified sample ($\text{sim}(x_i,x_i^*)$), as well as the change in the victim's output ($\text{diff}(f(x_i),f(x_i^*))$). }}
\label{fig:architecture}
\end{figure}

Adversarial example generation is a task aimed at testing the robustness of ML models, known as \textit{victims} in this context. The goal is to find small modifications to the input data that will change the model output even though the original meaning is preserved and the correct response remains the same. If such changed instances, known as adversarial examples, could be systematically found, it means the victim classifier is vulnerable to the attack and not robust.

In the context of classification, this setup (illustrated in Figure \ref{fig:architecture}) could be formalised through the following:
\begin{itemize}
    \item A training set $X_{train}$ and an attack set $X_{attack}$, each containing instances $(x_i, y_i)$, coupling the $i$-th instance features $x_i$ with its true class $y_i$,
    \item A victim model $f$, predicting a class label $\hat{y_i}$ based on instance features: $\hat{y_i}=f(x_i)$,
    \item A modification function (attack model) $m$, turning $x_i$ into an adversarial example $x^*_i=m(x_i)$.
\end{itemize}
Throughout this study, we use $y_i=1$ (positive class) to denote non-credible information and $0$ for credible content.

The goal of the attacker is to come up with the $m$ function. This process typically involves generating numerous variations of $x_i$ and querying the model's response to them until the best candidate is selected. An evaluation procedure assesses the success of the attack on the set $X_{attack}$ by comparing $x_i$ to $x^*_i$ (which should be maximally similar) and $f(x_i)$ to $f(x^*_i)$ (which should be maximally different).

{Consider the following real example observed in our evaluation:}
\begin{enumerate}
    \item {Within the propaganda recognition task, one of the instances in $X_{attack}$ contains a text fragment $x_i=$'\textit{Despite the hysteria of the left , it is impossible to see the Trump administration as anything but firm in its dealing with Russia.}', labelled as $y_i=1$ (propaganda technique used).}
    \item {The victim classifier (BiLSTM) correctly assigns the label $f(x_i)=1$ with 94.76\% certainty.}
    \item {An attacker (BERT-ATTACK) tests 26 different reformulations of the text, until it comes up with the modified version: $x^*_i=m(x_i)=$'\textit{\textbf{Given} the hysteria of the left , it is impossible to see the Trump administration as anything but firm in its dealing with Russia.}'}
    \item {The victim classifier changes its decision after the modification, assigning $f(x^*_i)=0$ (no propaganda) with 54.65\% certainty.}
    \item {This example is considered a good-quality AE, since it achieves a change in the classifier's decision ($f(x_i)\neq f(x^*_i)$) with a small change in text meaning.}
\end{enumerate}

\section{BODEGA tasks}
\label{sec:tasks}

In BODEGA we include four misinformation detection tasks:
\begin{itemize}
    \item {Hyperpartisan news (HN)},
    \item {Propaganda recognition (PR)},
    \item Fact checking (FC),
    \item Rumour detection (RD).
\end{itemize}

For each of these problems, we rely on an already established dataset with credibility labels provided by expert annotators. The tasks are all presented as text classification.

Whenever data split is released with a corpus, the training subset is included as $X_{train}$ -- otherwise we perform a random split. In order to enable the evaluation of AE generation solutions that carry a high computational cost, we define the $X_{attack}$ subset which is restricted to around 400 instances taken from the test set. The rest of the cases in the original test set are left out for future use as a development subset. Table \ref{tab:dataset} summarises the data obtained.

\begin{table}
    \caption{Four datasets used in BODEGA, with the task ID (see descriptions in text), number of instances in training, attack and development subsets, and an overall percentage of positive (non-credible) class.}
    \begin{tabular}{rrrrr}
    \hline
    \textbf{Task} & \textbf{Training} & \textbf{Attack} & \textbf{Dev.} & \textbf{Positive} \\
    \hline
        HN & 60,235 & 400 & 3,600 & 50.00\% \\
        PR & 12,675 & 416 & 3,320 & 29.42\% \\
        FC & 172,763 & 405 & 19,010 & 51.27\% \\
        RD & 8,694 & 415 & 2,070 & 32.68\% \\
    \hline
    \end{tabular}
    \label{tab:dataset}
\end{table}

Table \ref{tab:tasks} includes some examples of the credible and non-credible content in each task. {We can see how the non-credible examples often focus on particularly politically charged topics, trying to provoke an emotional reaction in readers. This a well-known aspect of misinformation }\citep{Bakir2017,Allcott2017}. In the following subsections, we outline the motivation, origin and data processing within each of the tasks. 

\begin{table*}
\caption{Examples of credible and non-credible content in each of the tasks: hyperpartisan news (HN), propaganda recognition (PR), fact checking (FC) and rumour detection (RD). See main text for references to data sources and labelling criteria.}
\small
    \centering
    \begin{tabular}{p{0.05\linewidth}p{0.42\linewidth}p{0.42\linewidth}}
    \hline
    \textbf{Task} & \textbf{Credible example} & \textbf{Non-credible example}\\
    \hline
        HN & Syria blamed for missed deadline on chemical arsenal\newline U.S. officials conceded that a Tuesday deadline for ridding Syria of hundreds of tons of liquid poisons would not be met, citing stalled progress in transporting the chemicals across war-ravaged countryside to ships that will carry them out of the region. But the officials insisted that the overall effort to destroy President Bashar Assad’s chemical arsenal was on track. "We continue to make progress, which has been the important part," State Department spokeswoman Marie Harf told reporters. "It was always an ambitious timeline, but we are still operating on the June 30th timeline for the complete destruction." (...) & Fox's Cavuto And Stein Try To Conflate 'Grubergate' With Vietnam And The Pentagon Papers \newline Over at Faux "news" this Tuesday, rather than focus on the newly released Senate torture report, it's been all Jonathan Gruber and "Grubergate" all the time and wall to wall coverage of another one of Darrell Issa's Obamacare witch hunts, otherwise known as a House Oversight Committee hearing. \newline As soon as I heard the hearing was scheduled I knew that it meant things were going to get ugly over at Fox, but not even in my wildest imagination could I have come up with this big giant turd that Neil Cavuto and his buddy Ben Stein managed to toss against the wall to attack Obamacare and Gruber. (...) \\ 
        PR & Leading Democratic senators like Robert Menendez, Ben Cardin and Chuck Schumer, who opposed Obama’s Iran deal may now feel that as opponents of the Trump administration, they are required to oppose any change to the Iran Nuclear Agreement Review Act. & What outcome would justify another U.S. war in a region where all the previous wars in this century have left us bleeding, bankrupt, divided and disillusioned? \\ 
        FC & \textit{\underline{Cersei Lannister.} She subsequently appeared in A Clash of Kings (1998) and A Storm of Swords (2000). \underline{A Clash of Kings.} A Clash of Kings is the second novel in A Song of Ice and Fire, an epic fantasy series by American author George R. R. Martin expected to consist of seven volumes. \newline $\rightarrow$ Cersei Lannister appears in a series that was written by an author from the United States.} & \textit{\underline{David Bowie.} During his lifetime, his record sales, estimated at 140 million worldwide, made him one of the world's best-selling music artists. \newline$\rightarrow$	David Bowie only sold records in Jamaica.} \\ 
        RD & BREAKING: Three gunmen involved in attack on Charlie Hebdo magazine, French Interior Minister Bernard Cazeneuve says. http://t.co/ak9mTVfJdR \newline @cnni the Islamic leaders should do something about the image of islam by speaking out against the terrorists \newline @cnni expel muslims from european soil and destroy all the mosques. \newline @cnni it's not the religion.  But how the people  interpret the writings and that's what causes them to do bad things. \newline @cnni terrorism needs concerted efforts from every citizen to fight it,religion is going beyond boundaries if it can cause terror attacks & Reports: \#CharlieHebdo suspects killed http://t.co/rsl4203bcQ \newline Damn, this is like a movie RT @HuffingtonPost Reports: \#CharlieHebdo suspects killed http://t.co/zCuZD1cure \newline ?@HuffingtonPost: Reports: \#CharlieHebdo suspects killed http://t.co/mWCSjh3CkH? superb simultaneous response by the French tactics unit. \newline @HuffingtonPost great news! No trial, no taxpayer money spent to support them. \newline @HuffingtonPost Good news !!! Alah Akbar !!    \newline @HuffingtonPost damnit!!! That's what those fuckers wanted!! Now they will be hailed as martyrs\textbf{....} \newline @HuffingtonPost Can you confirm the reports that those suspects were killed by French police? \newline (...) \\ 
    \hline
    \end{tabular}
    \label{tab:tasks}
\end{table*}

\subsection{{HN: Hyperpartisan news}}

Solutions for news credibility assessment, sometimes equated with \textit{fake news} detection, usually rely on one of three factors: (1) writing style \citep{Horne2017,Przybya2020}, (2) veracity of included claims \citep{Vlachos2014,Graves2018} or (3) context of social and traditional media  \citep{Shu2019,Liu2020a}.

In this task, we focus on the writing style. This means a whole news article is provided to a classifier, which has no ability to check facts against external sources, but has been trained on enough articles to recognise stylistic cues. The training data include numerous articles coming from sources with known credibility, allowing one to learn writing styles typical for credible and non-credible outlets.

In BODEGA, we employ a corpus of news articles \citep{potthast-etal-2018-stylometric} used for the task of \textit{Hyperpartisan News Detection} at SemEval-2019 \citep{kiesel-etal-2019-semeval}. The credibility was assigned based on the overall bias of the source, assessed by journalists from \textit{BuzzFeed} and \textit{MediaBiasFactCheck.com}\footnote{\url{https://zenodo.org/record/1489920}}. We use 1/10th of the training set (60,235 articles) and assign label $1$ (non-credible) to articles from sources annotated as hyperpartisan, both right- and left-wing.

See the first row of table \ref{tab:tasks} for examples: credible from \textit{Albuquerque journal}\footnote{\url{https://abqjournal.com/328734/syria-blamed-for-missed-deadline-on-weapons.html}} and non-credible from \textit{Crooks and Liars}\footnote{\url{http://crooksandliars.com/2014/12/foxs-cavuto-and-stein-try-conflate}}.

\subsection{{PR: Propaganda recognition}}

The task of propaganda recognition involves detecting text passages, whose author tries to influence the reader by means other than objective presentation of the facts, for example by appealing to emotions or exploiting common fallacies \citep{Smith1989a}. The usage of propaganda techniques does not necessarily imply falsehood, but in the context of journalism is associated with manipulative, dishonest and hyperpartisan writing. In BODEGA, we use the corpus accompanying SemEval 2020 Task 11 (\textit{Detection of Propaganda Techniques in News Articles}), with 14 propaganda techniques annotated in 371 newspapers articles by professional annotators \citep{Martino2020a}.

Propaganda recognition is a fine-grained task, with SemEval data annotated on the token level, akin to a Named Entity Recognition (NER) task. In order to cast it as a text classification problem as others here, we split the text on sentence level and assign target label equal 1 to sentences overlapping with any propaganda instances and 0 to the rest. Because only the training subset is made publicly available\footnote{\url{https://zenodo.org/record/3952415}}, we randomly extract 20 per cent  of documents for attack and development subsets.

See the second row of table \ref{tab:tasks} for examples -- the credible fragment with no propaganda technique and the non-credible, annotated as including flag-waving.

\subsection{FC: Fact checking}

Fact checking is the most advanced way human experts can verify credibility of a given text: by assessing the veracity of the claims it includes with respect to a knowledge base (drawing from memory, reliable sources and common sense). Implementing this workflow in AI systems as computational fact checking \citep{Graves2018} is a promising direction for credibility assessment. However, it involves many challenges -- choosing check-worthy statements \citep{Nakov2022}, finding reliable sources \citep{9891941}, extracting relevant passages \citep{karpukhin-etal-2020-dense} etc. Here we focus on the claim verification stage. The input of the task is a pair of texts – target claim and relevant evidence – and the output label indicates whether the evidence supports the claim or refutes it. It essentially is Natural Language Inference (NLI) \citep{MacCartney2009} in the domain of encyclopaedic knowledge and newsworthy events.

We use the data\footnote{\url{https://fever.ai/dataset/fever.html}} from FEVER shared task \citep{Thorne2018}, aimed to evaluate fact-checking solutions through a manually created set of evidence-claim pairs. Each pair connects a one-sentence claim with a set of sentences from Wikipedia articles, including a label of SUPPORTS (the evidence justifies the claim), REFUTES (the evidence demonstrates the claim to be false) or NOT ENOUGH INFO (the evidence is not sufficient to verify the claim). For the purpose of BODEGA, we take the claims from the first two categories\footnote{NOT ENOUGH INFO was excluded to cast the task as binary classification, in line with the other ones.}, concatenating all the evidence text\footnote{Including the titles, which are often an essential part of the context in case of encyclopaedic articles.}. The labels for the test set are not openly available, so we use the development set in this role.

See the examples in the third row of table \ref{tab:tasks}: the credible instance, where  combined evidence from two articles (titles underlined) supports the claim (after the arrow); and non-credible one, where the evidence refutes the claim.

\subsection{RD: Rumour detection}

A rumour is an information spreading between people despite not having a reliable source. In the online misinformation context, the term is used to refer to content shared between users of social media that comes from an unreliable origin, e.g. an anonymous account. Not every rumour is untrue as some of them can be later confirmed by established sources. Rumours can be detected by a variety of signals \citep{Al-Sarem2019}, but here we focus on the textual content of the original post and follow-ups from other social media users.

In BODEGA we use the Augmented dataset of rumours and non-rumours for rumour detection \citep{Han2019}, created from Twitter threads relevant to six real-world events (2013 Boston marathon bombings, 2014 Ottawa shooting, 2014 Sydney siege, 2015 Charlie Hebdo Attack, 2014 Ferguson unrest, 2015 Germanwings plane crash). The authors of the dataset started with the core threads annotated manually as rumours and non-rumours, then automatically augmented them with other threads based on textual similarity. We followed this by converting each thread to a flat feed of concatenated text fragments, including the initial post and subsequent responses. We set aside one of the events (Charlie Hebdo attack) for attack and development subsets, while others are included in the training subset.

See the last row of table \ref{tab:tasks} for examples, both regarding the \textit{Charlie Hebdo} shooting, but only the credible one is based on information from a credible source.

\section{Attack scenario}
\label{sec:attackscenario}
The adversarial attack scenarios are often classified according to what information is available to the attacker. The \textit{black-box} scenarios assume that no information is given on the inner workings of the targeted model and only system outputs for a given input can be observed. In \textit{white-box} scenarios, the model is openly available to the attacker, allowing them to observe its internal structure and understand how predictions are made. 

We argue neither of these scenarios is realistic in the practical misinformation detection setting, e.g. a content filter deployed in a social network. We cannot assume a model is available to the attacker since such information is usually not shared publicly; moreover, the model likely gets updated often to keep up with the current topics. On the other hand, the black-box scenario is too restrictive, as it assumes no information about the model is ever revealed. Also, once a certain design approach is popularised as the best performing in the NLP community, it tends to be applied to very many, if not most, solutions to related problems \citep{Church2022} -- this is especially noticeable in case of large language models, such as BERT \citep{Devlin2018} or GPT \citep{GPT2} and their successors.

For these reasons, in BODEGA we use the \textit{grey-box} approach. The following information is considered available to an attacker preparing AEs:
\begin{itemize}
    \item A ``hidden'' classifier $f$ that for any arbitrary input returns $f(x) \in \{0, 1\}$ and a likelihood score $s_f(x)$, i.e. a numerical representation on how likely a given example $x$ is to be assigned a positive class. This information is more helpful to attackers than only $f(x)$, which is typically set by applying a threshold $t_f$, e.g. $f(x) = 1 \iff s_f(x)> t_f$. {The threshold expresses the minimum value of the score necessary for the classifier to assign a positive label to the instance. Typically, this value is set to 0.5.}
    \item The general description of an architecture of classifier $f$, e.g. ‘a BERT encoder followed by a dense layer and softmax normalisation’.
    \item The training $X_{train}$, the development $X_{dev}$, and the evaluation $X_{attack}$ subsets.
\end{itemize}
This setup allows users of BODEGA to exploit weaknesses of classifiers without using the complete knowledge of the model, while maintaining some resemblance of practical scenarios.

{Note that the grey-box setup is significantly more challenging to attack compared the white-box scenario. In the latter, the attacker can directly see how the input features affect the output decision and modify those with the highest influence. Mathematically, this approach can be expressed in term of computing a gradient of the decision variable and following it -- thus the gradient-based methods} \citep{Zhang2020b}{. However, this is not possible to do in grey-box approach, where internal model weights, necessary for such procedure, are not revealed.}

Another choice that needs to be made concerns the goal of the attacker. Generally, adversarial actions are divided into \textit{untargeted} attacks, where any change of the victim's predictions is considered a success and \textit{targeted} attacks, which seek to obtain a specific response, aligned with the attacker's goals \citep{Zhang2020b}.

Consider a classifier $f$ that for a given instance $x_i$, with true value $y_i$, outputs class $f(x_i)$, which may be correct or incorrect. An \textit{untargeted} attack involves perturbing $x_i$ into $x^*_i$, such that $f(x_i)\neq f(x^*_i)$. A successful attack would undoubtedly show the brittleness of the classifier, but may not be necessarily helpful for a malicious user, e.g. if $y_i$ corresponded to malicious content, but the original response $f(x_i)$ was incorrect.

Taking into account the misinformation scenario, we consider the \textit{targeted} attack to satisfy the following criteria:
\begin{itemize}
    \item The true class corresponds to non-credible content, i.e. $y_i=1$,
    \item The original classifier response was correct, i.e. $f(x_i)=y_i$.
\end{itemize}
Success in this attack corresponds to a scenario of the attacker preparing a piece of non-credible content that is falsely recognised as credible thanks to the adversarial modification. We therefore use only a portion of the evaluation $X_{attack}$ subset for this kind of attack. 

By \textit{non-credible content} we mean:
\begin{itemize}
    \item In case of {hyperpartisan news}, an article from a hyperpartisan source,
    \item In case of {propaganda recognition}, a sentence with a propaganda technique,
    \item In case of fact checking, a statement refuted by the provided evidence,
    \item In case of rumour detection, a message feed starting from a post including a rumour.
\end{itemize}

In BODEGA, both untargeted and targeted attacks can be evaluated. 

All of the text forming an instance can be modified to make {an adversarial attack}. In case of fact checking, this includes both the claim and the evidence. Similarly for rumour detection, not only the original rumour, but also any of the follow-up messages in the thread {are included in the text instance}. This corresponds to the real-life scenario, where all of the above content is user generated and can to some degree be influenced by an attacker (see further discussion on this matter in section \ref{sec:realitycheck}).

Finally, note that BODEGA imposes no restriction on the number of queries sent to the victim, i.e. the number of variants an attacker is allowed to test for each instance before providing the final modification. This number would typically be limited, especially in a security-oriented application \citep{Chen2022}. {However, the constraints might be very different depending on a particular application scenarios. Some services might impose very strict limits on a number submissions a client can make within a specified time, while others might allow many more attempts. If an attacker knows the data the victim classifier was trained on, they can even train a surrogate classifier and issue as many queries as needed. Thus, in order to provide a comprehensive evaluation, the number of queries is not limited in BODEGA, but it is recorded as an evaluation metric (see the next section)}. 

\section{Evaluation}
\label{sec:evaluation}

Preparing adversarial examples involves balancing two goals in the adversarial attack (see Figure \ref{fig:architecture}):
\begin{enumerate}
    \item Maximising $\text{diff}(f(x_i), f(x^*_i))$ – difference between the classes predicted by the classifier for the original and perturbed instance,
    \item Maximising $\text{sim}(x_i, x^*_i)$ – similarity between the original and perturbed instance.
\end{enumerate}

If (1) is too small, the attack has failed, since the classifier preserved the correct prediction. If (2) is too small, the attack has failed, since the necessary perturbation was so large it defeated the original purpose of the text.

This makes the evaluation multi-criterion and challenging since neither of these factors measured in isolation reflects the quality of AEs. The conundrum is usually resolved by setting the minimum similarity (2) to a fixed threshold (known as \textit{perturbation constraint}) and measuring the reduction in classification performance, i.e. accuracy reduction \citep{Zhang2020b}. This can be problematic as there are no easy ways to decide the value of the threshold that will guarantee that the class remains valid. The issue is especially relevant for a task as subtle as credibility analysis --  e.g. how many word swaps can we do on a real news piece before it loses credibility?

In BODEGA we avoid this problem by inverting the approach. Instead of imposing constraints on goal (2) and using (1) as evaluation measure, we impose constraints on (1) and use (2) for evaluation. {Specifically, we only count the instances when the modification was sufficient to change the classifier's decision (1), and treat text similarity (2) as quality evaluation measure.}


We define an adversarial modification quality score, called \textit{BODEGA score}. BODEGA score always lies within 0-1 and a high value indicates good quality modification preserving the original meaning (with score=1 corresponding to no visible change), while low value indicates poor modification, altering the meaning (with score=0 corresponding to completely different text).

In the remainder of this section, we discuss the similarity measurement techniques we employ and outline how they are combined to form a final measure of attack success.

\subsection{Semantic score}

The first element used to measure meaning preservation is based on \textit{BLEURT} \citep{sellam-etal-2020-bleurt}. BLEURT was designed to compute the similarity between a candidate and reference sentences in evaluating solutions for natural language generation tasks (e.g. machine translation). The underlying model is trained to return values between 1 (identical text) and 0 (no similarity).

BLEURT helps to properly assess semantic similarity; for example, replacing a single word with its close synonym will yield high score value, while using a completely different one will not. However, BLEURT is trained to interpret multi-word modifications (i.e. paraphrases) as well, leading to better correlation with human judgement than other popular measures, e.g. BLEU or BERTScore. This is possible thanks to fine-tuning using synthetic data covering various types of semantic differences, e.g. \textit{contradiction} as understood in the NLI (Natural Language Inference) task. This is especially important for our usecase, helping to properly handle the situations where otherwise small modifications completely change the meaning of text (e.g. a negation), rendering an AE unusable.

In BODEGA, we use the pyTorch implementation of BLEURT\footnote{\url{https://github.com/lucadiliello/bleurt-pytorch}}, choosing the recommended\footnote{\url{https://github.com/google-research/bleurt}} \texttt{BLEURT-20} variant. Since the score is only \textit{calibrated} to the 0-1 range, other numbers can be produced as well. Thus, our semantic score is equal to BLEURT  (clipped to 0-1 if necessary). Finally, since BLEURT is a sentence-level measure and our tasks involve longer text fragments\footnote{Except propaganda detection, where input is a single sentence.}, we (1) split the text into sentences\footnote{Except fact checking, where we simply split evidence from claim.} using LAMBO \citep{Przybya2022}, (2) find the pairs of sentences sentences from the original and modified text that are most similar using Levenshtein distance and (3) compute semantic similarities between sentence pairs, returning its average as semantic score.

\subsection{Character score}

Levenshtein distance is used to express how different one string of characters is from another. Specifically, it computes the minimum number of elementary modifications (character additions, removals, replacements) it would take to transform one sequence into another \citep{1965}.

Levenshtein is a simple measure that does not take into account the meaning of the words. However, it is helpful to properly assess modifications that rely on graphical resemblance. For example, one family of adversarial attacks relies on replacing individual characters in text (e.g. \textit{call} to \textit{ca$||$}), altering the attacked classifier’s output. The low value of Levenshtein distance in this case represents the fact that such modification may be imperceptible for a human reader.

In order to turn Levenshtein distance $lev\_dist(a,b)$ into a character similarity score,  we compute the following:
\[
\text{Char\_score}(a, b) = 1 - \frac{lev\_dist(a, b)}{max(|a|, |b|)}
\]
$\text{Char\_score}$ is between 0 and 1, with higher values corresponding to larger similarity, with $\text{Char\_score}(a, b) = 1$  if $a$ and $b$ are the same and $\text{Char\_score}(a, b) = 0$ if they have no common characters at all.

\subsection{BODEGA score}

The BODEGA score for a pair of original text $x_i$ and modified text $x^*_i$ is defined as follows:
\begin{align*}
\text{BODEGA\_score}(x_i,x^*_i)& = \text{Con\_score}(x_i,x^*_i)  \times\\
     \text{Sem\_score}&(x_i,x^*_i)  \times \text{Char\_score}(x_i,x^*_i),
\end{align*}
where $\text{Sem\_score}(x_i,x^*_i)$ is semantic score; $\text{Char\_score}(x_i,x^*_i)$ is character score; and $\text{Con\_score}(x_i,x^*_i)$ is confusion score, which takes value $1$ when an adversarial example is produced and succeeds in changing the victim's decision (i.e. $f(x_i)\neq f(x^*_i)$) and $0$ otherwise.

The overall attack success measure is computed as an average over BODEGA scores for all instances in the attack set available in a given scenario (targeted or untargeted). The success measure reaches 0 when the AEs bear no similarity to the originals, or they were not created at all. The value of 1 corresponds to the situation, unachievable in practice, when AEs change the victim model's output with immeasurably small perturbation.


Many adversarial attack methods include tokenisation that does not preserve the word case or spacing between them. Our implementation of the scoring disregards such discrepancies between input and output, as they are not part of the intended adversarial modifications.

Apart from BODEGA score, expressing the overall success, the intermediate measures can paint a fuller picture of the strengths and weaknesses of a particular solution:
\begin{itemize}
    \item Confusion score – in how many of the test cases the victim's decision was changed,
    \item Semantic score – an average over the cases with changed decision,
    \item Character score – an average over the cases with changed decision.
\end{itemize}

We also report the number of queries made to the victim, averaged over all instances.

\section{Victim classifiers}
\label{sec:victimclassifiers}

A victim classifier is necessary to perform an evaluation of an AE generation solution. We include implementations of text classifier based on various common architectures: a recurrent neural network (BiLSTM) trained from scratch; and fine-tuned language models: small masked model (BERT), large generative model (GEMMA2B) and a very large generative model (GEMMA7B), delivering state-of-the-art results in the established benchmarks.

This component of BODEGA could be easily replaced by newer implementations, either to test a robustness of a specific classifier architecture, or to have a better understanding of applicability of a given AE generation solution.

\subsection{BiLSTM}

The recurrent network is implemented using the following layers:
\begin{itemize}
    \item An embedding layer, representing each token as vector of length 32,
    \item Two LSTM \citep{Hochreiter1997a} layers (forwards and backwards), using hidden representation of length 128, returned from the edge cells and concatenated as document representation of length 256,
    \item A dense linear layer, computing two scores representing the two classes, normalised to probabilities through softmax.
\end{itemize}

The input is tokenised using BERT uncased tokeniser (see below). The maximum allowed input length is 512, with padding as necessary. For each of the tasks, a model instance is trained from scratch for 10 epochs by using Adam optimiser \citep{Kingma2015a}, a learning rate of 0.001 and batches of 32 examples each. The implementation uses \textit{PyTorch}.

\subsection{BERT}

As a baseline pretrained language model, we use BERT in the base variant \citep{Devlin2018}. The model is fine-tuned for sequence classification using Adam optimiser with linear weight decay \citep{Loshchilov}, starting from 0.00005, for 5 epochs. We use maximum input length of 512 characters and a batch size of 16. The training is implemented using the \textit{Hugging Face Transformers} library \citep{wolf-etal-2020-transformers} (\texttt{bert-base-uncased} model).

\subsection{Gemma}
In order to assess the vulnerability of the large language models to AEs, we include \textit{Gemma} \citep{GemmaTeam2024}. Gemma is a recent generative language model, derived from Google's \textit{Gemini} models and following the same design principles as the GPT family \citep{GPT2}. We include both the smaller variant with 2 billion parameters, as well as the full 7-billion model, loaded through \textit{Hugging Face Transformers}. They have been evaluated in multiple benchmarks and the latter has shown the best performance among the openly available large language models \citep{GemmaTeam2024}. 

The fine-tuning was performed using the same procedure as for BERT. However, in order to keep the computing requirements under control, we applied parameter-efficient fine-tuning \citep{lialin2023scaling}. Namely, we used QLoRA optimisation \citep{NEURIPS2023_1feb8787}, based on of Low Rank Adaptation (LoRA) \citep{hu2021lora} with reduced numerical precision. These are implemented using the \textit{Hugging Face}'s libraries \texttt{peft} and \texttt{bitsandbytes}, respectively. 

\section{AE generation solutions}
\label{sec:baselinesolutions}
Within BODEGA, we include the AE generation solutions implemented in the \textit{OpenAttack} framework. We exclude the approaches for white-box scenario (gradient-based) and those that yielded poor performance in preliminary tests. We test 8 approaches:
\begin{itemize}
    \item \textbf{BAE} \citep{garg-ramakrishnan-2020-bae} uses BERT \citep{Devlin2018} as a masked language model to generate word candidates that are likely in a given context. This includes both replacing existing tokens as well as inserting new ones.
    \item \textbf{BERT-ATTACK} \citep{li-etal-2020-bert-attack} is a very similar approach, which starts with finding out if a word is vulnerable by checking victim's response to its masking. The chosen words are replaced using BERT candidates, but unlike in BAE, no new words are inserted.
    \item \textbf{DeepWordBug} \citep{Gao2018a} works at the character level, seeking modifications that are barely perceptible for humans, but will modify an important word into one unknown to the attacked model. The options include character substitutions, removal, insertion and reordering.
    \item \textbf{Genetic} \citep{alzantot-etal-2018-generating} is using the genetic algorithm framework. A \textit{population} includes variants of text built by word replacements (using GloVe representation to ensure meaning preservation), the most promising of which can replicate and combine until a successful AE is found.
    \item \textbf{SememePSO} \citep{zang-etal-2020-word} employs a related framework, namely Particle Swarm Optimisation (PSO). A group of \textit{particles}, each representing a text modification with a certain probability of further changes (\textit{velocity}), moves through the feature space until an optimal position is found.
    \item \textbf{PWWS} \citep{ren-etal-2019-generating} is a classical greedy word replacement approach. However, it differs from the majority of the solutions by using \textit{WordNet}, instead of vector representations, to obtain synonym candidates.
    \item \textbf{SCPN} \citep{Iyyer2018} performs paraphrasing of the whole text through a bespoke encoder-decoder model. In order to train this model, the authors generate a dataset of paraphrases through backtranslation from English to Czech.
    \item \textbf{TextFooler} \citep{DBLP:conf/aaai/JinJZS20} is a greedy word-substitution solution. Unlike other similar approaches, it takes into account the syntax of the attacked text, making sure the replacement is a valid word that agrees with the original regarding its part of speech. This help to make sure the AE is fluent and grammatically correct.
\end{itemize}

The main problem the presented solutions try to solve is essentially maximising a goal function (victim's decision) in a vast space of possible modifications to input text, which is further complicated by its discrete nature. Direct optimisation is not computationally feasible here, giving way to methods that are greedy (performing the change that improves the goal the most) or maintain a population of varied candidate solutions (PSO and evolutionary algorithms). The majority of the solutions operate on word-level, seeking replacements that would influence the classification result without modifying the meaning. The exceptions are sentence-level SCPN, performing paraphrasing of entire sentences, and character-level DeepWordBug, replacing individual characters in text to preserve superficial similarity to the original. They all use victims' scores to look for most promising modifications, except for SCPN, which operates blindly, simply generating numerous possible paraphrases. 

All of the attackers are executed with their default functionality, except for BERT-ATTACK, that we use without the generation of subword permutations, which is prohibitively slow for longer documents. Just like the victim classifier, the AE solution interface in BODEGA allows for new solutions to be added and tested as the field progresses.

\subsection{Classification performance}

\begin{table}
    \caption{Performance of the victim classifiers, expressed as F-score over the attack subset.}
    \begin{tabular}{rrrrr}
    \hline
    & \textbf{BiLSTM} & \textbf{BERT} & \textbf{GEMMA2B} & \textbf{GEMMA7B}\\
    \hline
        HN & 0.7076 & 0.7544 & \textbf{0.7792} & 0.7603 \\
        PR & 0.4857 & 0.6410 & 0.6271 & \textbf{0.6840}\\
        FC & 0.7532 & 0.9360 & 0.9701 & \textbf{0.9727} \\
        RD & 0.6234 & 0.7547 & \textbf{0.7609} & 0.7229 \\
    \hline
        \textbf{Parameters} & 1M & 340M & 2B & 7B \\
        \hline
    \end{tabular}
    \label{tab:victims}
\end{table}

Table \ref{tab:victims} shows the performance of the victim classifiers, computed as F-score over the test data (combined development and attack subsets). As expected, BERT easily outperforms a neural network trained from scratch. The credibility assessment tasks are subtle and the amount of data available for training severely limits the performance. Thus, the BERT model has an advantage by relying on knowledge gathered during pretraining. This is demonstrated by the performance gap being the largest for the dataset with the least data available (propaganda detection) and the smallest for the most abundant corpus (hyperpartisan news). The Gemma models perform even better than BERT in all tasks. However, the improvement is not as spectacular (a few percent) and GEMMA7B does not provide uniformly better results than the 2-billion model.

\section{Experiments}
\label{sec:results}

{The purpose of the experiments is to test the BODEGA framework in action and improve our understanding of the vulnerability of content filtering solutions to adversarial actions. This will also establish a baseline for systematic evaluation of future classifiers and AE generators.} To that end, we test the attack performance for:
\begin{itemize}
    \item four tasks (HN, PR, FC, RD),
    \item eight attackers (BAE, BERT-ATTACK, DeepWordBug, Genetic, SememePSO, PWWS, SCPN, textFooler),
    \item four victims (BiLSTM, BERT, GEMMA2B, GEMMA7B),
    \item two scenarios (untargeted and targeted).
\end{itemize}
In total, $4 \times 8 \times 4 \times 2 = 256$ experiments are performed, each evaluated using the measures introduced in section \ref{sec:evaluation}.

The full results are shown in the appendix. Here we present an analysis focused on key questions:
\begin{itemize}
    \item Q1: Which attack method delivers the best performance?
    \item Q2: Are the modern large language models less vulnerable to attacks than their predecessors?
    \item Q3: How many queries are needed to find adversarial examples?
    \item Q4: Does targeting make a difference in attack difficulty?
\end{itemize}
Moreover, we perform a manual analysis of the most promising AEs (section \ref{sec:manual}).

\subsection{Q1: Attack methods}
\label{sec:attacks}

\begin{table}
    \caption{The results of adversarial attacks, averaged over all victim classifiers, in four misinformation detection tasks (untargeted). Evaluation measures include BODEGA score, confusion score, semantic score, character score and number of queries to the attacked model per example. The best score in each task is in boldface.}
    \small
    \begin{tabular}{rrrrrrr}
    \hline
    \textbf{Task} & \textbf{Method} & \textbf{BODEGA} & \textbf{Confusion} & \textbf{Semantic} & \textbf{Character} & \textbf{Queries}   \\
    \hline
        HN & BAE  & {0.36} & {0.60} & {0.61} & {0.97} & {589.39} \\
         & BERT-ATTACK  & \textbf{0.56} & \textbf{0.90} & {0.63} & {0.97} & {910.00} \\
         & DeepWordBug  & {0.25} & {0.33} & \textbf{0.78} & \textbf{1.00} & {390.95} \\
         & Genetic  & {0.38} & {0.81} & {0.48} & {0.98} & {1740.55} \\
         & SememePSO  & {0.19} & {0.40} & {0.50} & {0.99} & {309.10} \\
         & PWWS  & {0.37} & {0.79} & {0.48} & {0.98} & {2051.62} \\
         & SCPN  & {0.00} & {0.82} & {0.09} & {0.02} & {11.75} \\
         & TextFooler  & {0.34} & {0.77} & {0.46} & {0.96} & {792.91} \\
    \hline
        PR & BAE  & {0.14} & {0.21} & {0.71} & {0.94} & {33.31} \\
         & BERT-ATTACK  & {0.46} & {0.72} & {0.69} & {0.91} & {76.40} \\
         & DeepWordBug  & {0.20} & {0.26} & \textbf{0.79} & \textbf{0.96} & {27.33} \\
         & Genetic  & \textbf{0.49} & \textbf{0.84} & {0.65} & {0.89} & {886.55} \\
         & SememePSO  & {0.41} & {0.68} & {0.67} & {0.89} & {99.51} \\
         & PWWS  & {0.46} & {0.74} & {0.67} & {0.90} & {132.20} \\
         & SCPN  & {0.11} & {0.54} & {0.38} & {0.48} & {11.54} \\
         & TextFooler  & {0.41} & {0.72} & {0.65} & {0.87} & {62.26} \\
    \hline
        FC & BAE  & {0.35} & {0.53} & {0.69} & {0.96} & {78.58} \\
         & BERT-ATTACK  & \textbf{0.57} & \textbf{0.83} & {0.72} & {0.95} & {153.37} \\
         & DeepWordBug  & {0.26} & {0.31} & \textbf{0.83} & \textbf{0.98} & {54.10} \\
         & Genetic  & {0.52} & {0.77} & {0.70} & {0.95} & {846.25} \\
         & SememePSO  & {0.44} & {0.65} & {0.71} & {0.96} & {145.06} \\
         & PWWS  & {0.48} & {0.69} & {0.72} & {0.96} & {225.98} \\
         & SCPN  & {0.07} & {0.66} & {0.30} & {0.33} & {11.66} \\
         & TextFooler  & {0.46} & {0.70} & {0.70} & {0.94} & {109.77} \\
    \hline
        RD & BAE  & {0.10} & {0.24} & {0.42} & {0.98} & {310.71} \\
         & BERT-ATTACK  & \textbf{0.25} & \textbf{0.62} & {0.42} & {0.94} & {860.04} \\
         & DeepWordBug  & {0.13} & {0.19} & \textbf{0.70} & \textbf{0.99} & {235.85} \\
         & Genetic  & {0.24} & {0.53} & {0.46} & {0.96} & {2605.13} \\
         & SememePSO  & {0.12} & {0.26} & {0.47} & {0.97} & {330.20} \\
         & PWWS  & {0.21} & {0.46} & {0.46} & {0.96} & {1107.09} \\
         & SCPN  & {0.01} & {0.41} & {0.17} & {0.11} & {11.40} \\
         & TextFooler  & {0.18} & {0.46} & {0.44} & {0.92} & {654.20} \\
    \hline
    \end{tabular}
    \label{tab:methodresults}
\end{table}

Table \ref{tab:methodresults} compares the performance of the untargeted attack methods in various tasks, averaged over victim models.

The hyperpartisan news detection task is relatively easy for generating AEs. BERT-ATTACK achieves the best BODEGA score of 0.56, which is possible due to changing the decision on 90 per cent  of the instances while preserving high similarity, both in terms of semantics and characters. However, DeepWordBug (a character-level method) provides the best results in terms of semantic similarity, changing less than 1 per cent  of characters on average. The only drawback of this method is that it works in 25 per cent  of the cases, failing to change the victim's decision in the remaining ones.

The propaganda recognition task significantly differs from the previous task in terms of text length, including individual sentences rather than full articles. As a result, every word is more important and it becomes much harder to make the changes imperceptible, resulting in lower character similarity scores. This setup appears to favour the Genetic method, obtaining the best BODEGA score: 0.49. This approach performs well across the board, but it comes at a high cost in terms of model queries. Even for the short sentences in propaganda recognition, a victim model is queried over 800 times, compared to less than 150 for all other methods.

Fact checking resembles the propaganda recognition in terms of relatively short text fragments, but the best-performing method is BERT-ATTACK. As for hyperpartisan news, DeepWordBug achieves high similarity, but succeeds in finding an AE relatively rarely -- 26 per cent  of times.

Finally, the rumour detection task in the untargeted scenario appears to be the hardest problem to attack. Here the best methods reaches BODEGA score of 0.25, indicating low usability, mostly due to low confusion rates -- barely above 60 per cent . This may be because rumour threads consist of numerous posts, each having some indication on the credibility of the news, forcing an attacker to make many modifications to change the victim's decision. The text of Twitter messages is also far from regular language, making the challenge harder for methods using models pretrained on well-formed text (e.g. BERT-ATTACK). It has to be noted however that this setup is equally problematic to the meaning preservation measurement (semantic score), thus suggesting these results should be taken cautiously.

{Regarding the performance of the included attack methods, we can observe the following:}
\begin{itemize}
    \item {Approaches relying on local changes (e.g. BERT-ATTACK, DeepWordBug) work better than global rephrasers (SCPN), because they are able to deliver more candidates for AEs and thus have more chances for success.}
    \item {Character-replacing solutions (e.g. DeepWordBug) maintain high similarity, both in semantic and Levenstain measures, but suffer in terms of confusion rate. Clearly, sometimes changing a whole word is necessary to trigger a decision change.}
    \item {Methods relying on language models for meaning representation (esp. BERT-ATTACK) obtain better results than those relying on GloVe (Genetic) or WordNet (PWWS). This is likely because the older methods are not context-sensitive, resulting in less appropriate replacements, visible as reduced semantic scores.}
    \item {Solutions performing a very extensive search (esp. Genetic) find good AEs only for short text: propaganda and fact-checking. They become unfeasible for longer content, e.g. news.}
    \item {Even solutions with apparently similar designs (BAE and BERT-ATTACK) can deliver vastly different performance due to smaller details in their implementation.}
\end{itemize}

\subsection{Q2: Victim size and vulnerability}
\label{sec:victims}

\begin{figure}
\includegraphics[width=0.8\linewidth]{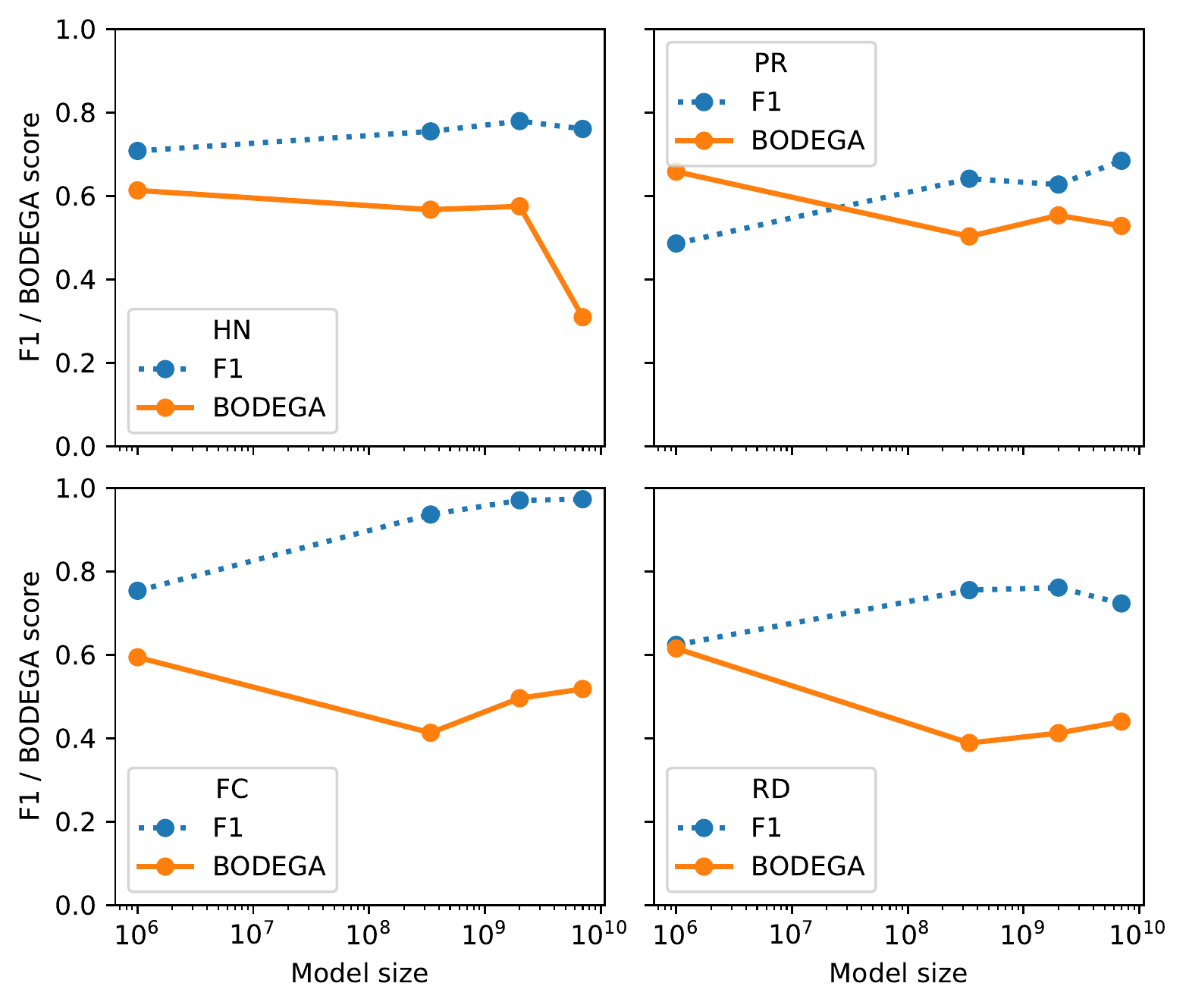} 
\vspace{0.5cm}
\caption{Classification performance (F1 score) and vulnerability to targeted attacks (BODEGA score) of models according to their size (parameter count, logarithmic scale), for different tasks.}
\label{fig:victims}
\end{figure}

Figure \ref{fig:victims} plots the performance and vulnerability to targeted attacks (BODEGA score of the most successful method) of models of increasing size: BiLSTM, BERT, GEMMA2B, GEMMA7B. We can see that while the classification scores almost universally improve with larger models (albeit with diminishing returns), the robustness assessment paints a more complex picture. 

BiLSTM, which is by far the smallest model, is also clearly the most vulnerable to attacks. However, the results for the large pretrained models are surprising: the smallest of them (BERT) appears to be the most robust, except for one task (HN). {This effect is the strongest for the FC task, where the best attacker on the GEMMA7B model achieves a score 27\% higher than in the attack against BERT.} For two of the tasks (FC and RD), this pattern holds even within the same model family, with the smaller GEMMA model showing lower vulnerability.

Overall, new and more accurate language models are not less vulnerable to attacks, as one would hope. In the application scenarios involving adversarial actors, such as credibility assessment, smaller solutions may thus be a more appropriate choice. {This observation is a contribution to the wider question of vulnerability of LLMs to a adversarial actions }\citep{YAO2024100211,Goto_2024}{. While this is a new research area, preliminary results are concordant with ours, namely showing larger models as not necessarily increasing robustness over the smaller predecessors} \citep{liu2024robustnesstimeunderstandingadversarial}. {Our results do not explain why the robustness does not increase with model size as classification performance does, and we leave this problem as an interesting question for future research.}

\subsection{Q3: Number of queries}
\label{sec:queries}

\begin{figure}
\includegraphics[width=\linewidth]{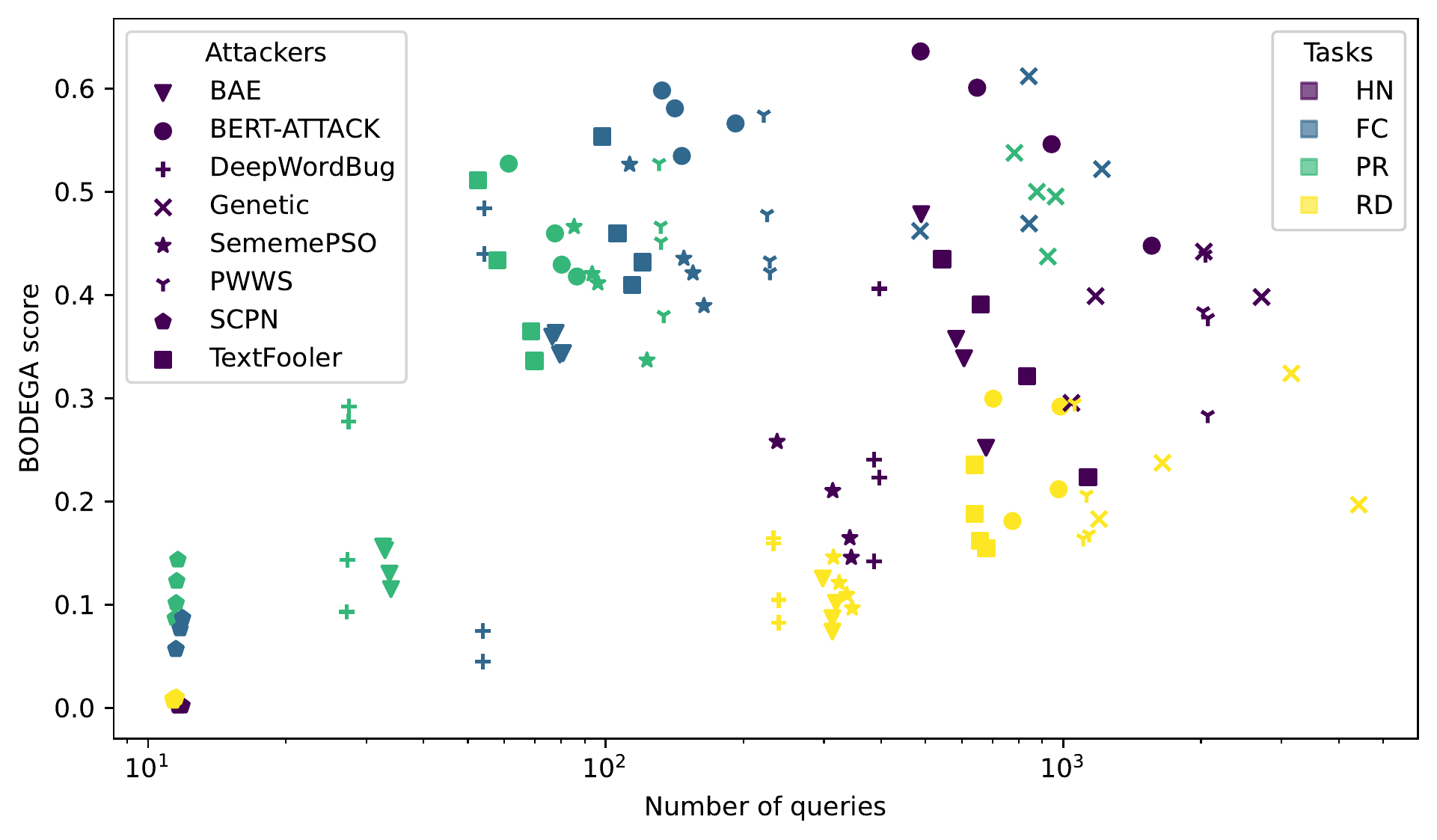}
\vspace{0.5cm}
\caption{Results of the targeted attacks (y axis, BODEGA score) plotted against the number of queries necessary (x axis, logarithmic) for various attack methods (symbols) and tasks (colours).}
\label{fig:queries}
\end{figure}

Figure \ref{fig:queries} illustrates the number of queries necessary to perform attacks with various levels of success. Primarily, we can see the results are grouped according to the task being attacked. The tasks involving long text (HN and RD) both require many queries: for each attacked example, from several hundred to several thousand attempts are needed to find an adversarial variant. These two tasks differ in terms of success, with hyperpartisan news obtaining some of the highest BODEGA scores and rumour detection: the lowest. The tasks involving shorter text (FC and PR) have similarly high success rate, but good attacks require much less queries: from just over 100 (FC) to less then 60 (PR).

In terms of attack methods, BERT-ATTACK clearly achieves the best BODEGA score for most tasks. However, it requires many queries -- even though not as many as the Genetic approach. Among the methods that work with less queries, often with little cost in terms of performance loss, we can distinguish TextFooler and DeepWordBug.

\subsection{Q4: Targeting}
\label{sec:targeting}

\begin{table}
\small
    \caption{A comparison of the results -- highest BODEGA score and corresponding number of queries -- in the untargeted (U) and targeted (T) scenario for various tasks and victims. The better values (higher BODEGA scores and lower number of queries) are highlighted.}
    \centering
    \begin{tabular}{rrrrrrrrrr}
    \hline
    & & \multicolumn{2}{c}{\textbf{BiLSTM}} & \multicolumn{2}{c}{\textbf{BERT}} & \multicolumn{2}{c}{\textbf{GEMMA2B}} & \multicolumn{2}{c}{\textbf{GEMMA7B}}\\
    & & \textbf{U} & \textbf{T} & \textbf{U} & \textbf{T} & \textbf{U} & \textbf{T} & \textbf{U} & \textbf{T}\\
\hline
        HN & B. score & \textbf{0.64} & {0.61} & \textbf{0.60} & {0.57} & {0.55} & \textbf{0.57} & \textbf{0.45} & {0.31}\\ 
         & Queries & \textbf{487.85} & {565.05} & \textbf{648.41} & {753.91} & {942.98} & \textbf{761.53} & \textbf{1560.76} & {2313.33}\\ 
 \hline
        PR & B. score & {0.54} & \textbf{0.66} & {0.50} & \textbf{0.50} & {0.50} & \textbf{0.55} & {0.44} & \textbf{0.53}\\ 
         & Queries & {782.15} & \textbf{50.14} & {962.40} & \textbf{99.95} & {876.06} & \textbf{94.05} & {925.58} & \textbf{110.32}\\ 
 \hline
        FC & B. score & \textbf{0.61} & {0.59} & \textbf{0.53} & {0.41} & \textbf{0.57} & {0.50} & \textbf{0.58} & {0.52}\\ 
         & Queries & {840.99} & \textbf{123.24} & \textbf{146.73} & {207.23} & \textbf{192.25} & {254.22} & \textbf{141.70} & {173.86}\\ 
 \hline
        RD & B. score & {0.32} & \textbf{0.62} & {0.20} & \textbf{0.39} & {0.30} & \textbf{0.41} & {0.21} & \textbf{0.44}\\ 
         & Queries & {3150.24} & \textbf{153.61} & {4425.11} & \textbf{174.03} & \textbf{703.07} & {1108.21} & {977.27} & \textbf{202.18}\\ 
 \hline
    \end{tabular}
    \label{tab:targetingresults}
\end{table}

Table \ref{tab:targetingresults} compares targeted and untargeted scenarios in terms of performance -- the best BODEGA score and the number of queries needed to achieve it. Interestingly, the individual score differences can be quite high, but the pattern depends on the classification task. The targeted task is always harder for news bias assessment (except BERT) and fact checking. The untargeted one is always much more challenging for propaganda recognition and rumour detection.

\subsection{Manual analysis}
\label{sec:manual}
In order to better understand how a successful attack might look like, we manually analyse some of them. This allows us observe what types of adversarial modifications are the weakest point of the classifier, as well as verify if attack success scoring using automatic measures is aligned with the human judgement.

For that purpose, we select 20 instances with the highest BODEGA score from the untargeted interactions between a relatively strong attacker (BERT-ATTACK) and a relatively weak victim (BiLSTM), within all tasks. Next, we label the AEs according to the degree they differ from the original text:\footnote{Note that while these categories might overlap, e.g. a typographic replacement significantly affecting the overall meaning, such cases were not encountered in practice during the analysis.}
\begin{enumerate}
    \item \textbf{Synonymous}: the text is identical in meaning to the original.
    \item \textbf{Typographic}: change of individual characters, e.g. resembling sloppy punctuation or typos, likely imperceptible.
    \item \textbf{Grammatical}: change of the syntax of the sentence, e.g. replacing a verb with a noun with the same root, possibly making the text grammatically incorrect,
    \item \textbf{Semantic-small}: changes affecting the overall meaning of the text, but to a limited degree, unlikely to affect the credibility label,
    \item \textbf{Semantic-large}: significant changes of the meaning of the text, indicating the original credibility may not apply,
    \item \textbf{Local}: changes of any degree higher than Synonymous, but present only in a few non-crucial sentences of a longer text, leaving others to carry the original meaning (applies to tasks with many sentences, i.e. RD and HN).
\end{enumerate}
The changes labelled as Semantic-large indicate attack failure, while others denote success with varying visibility of the modification.

\begin{table}
    \caption{Number of AEs using different modifications among the best 20 instances (according to BODEGA score) in each task, using BiLSTM as victim and BERT-ATTACK as attacker.}
    \begin{tabular}{rrrrrr}
    \hline
     & \multicolumn{4}{c}{\textbf{number of instances}} &\\
    \textbf{AE degree} & \textbf{HN} & \textbf{PR} & \textbf{FC} & \textbf{RD} &  \textbf{\% of all} \\
    \hline
        Synonymous & 6 & 10 & 2 & 5 &29\% \\
        Typographic & 0 & 5 & 8 & 0 &16\%\\
        Grammatical & 0 & 4 & 3 & 2 &11\%\\
        Semantic-small & 0 & 1 & 2 & 3 &7\%\\
        Local & 13 & - & - & 2 & 19\%\\
        \hline
        Semantic-large & 1 & 0 & 5 & 8 &17\%\\
    \hline
    \end{tabular}
    \label{tab:manual}
\end{table}

Table \ref{tab:manual} shows the quantitative results of the manual analysis, while table \ref{tab:examples} includes some examples. Generally, a large majority of these attacks (82.5 per cent ) were successful in maintaining the original meaning, confirming the high BODEGA score assigned to them. However, significant differences between the tasks are visible.

Consistently with the results of automatic analysis, rumour detection appears to be the most robust, resulting in many attacks changing the original meaning. Even though oftentimes only a word or two is changed, it affects the meaning of the whole Twitter thread, since the follow-up messages do not repeat the content, but often deviate from the topic (see EX4 in table \ref{tab:examples}). The opposite happens for hyperpartisan news: a singular change does not affect the overall message, as the news article are typically redundant and maintain their sentiment throughout (see EX6). As a result, the HR task is one of the most vulnerable to attacks.

It is also interesting to compare the two tasks with shorter text: fact checking and propaganda recognition. While the FC classifier shows a large vulnerability to typographic changes (esp. in punctuation, see EX2), many of the changes performed by the attackers affect important aspects of the content (e.g. names or numbers, see EX5), making the AE futile. The propaganda recognition, on the other hand, appears to rely on stylistic features, allowing the AE generation while preserving full synonymy (see EX1) or just introducing grammatical issues (see EX3).

\begin{table}
    \caption{Some examples of adversarial modifications that were successful (i.e. resulted in changed classifier decision), performed by BERT-ATTACK against BiLSTM, including identifier (mentions in text), task and type of modification. Changes are highlighted in boldface.}
\small
    \begin{tabular}{p{0.11\linewidth}p{0.40\linewidth}p{0.40\linewidth}}
    \hline
    \textbf{Id., task, type} & \textbf{Original example} & \textbf{Adversarial example}\\
    \hline
        EX1 PR \newline Synonymous & Puerto Rico's housing secretary, Fernando Gil, says the number of \textbf{homes} destroyed by the hurricane totals about 70,000 so far, and homes with major damage have amounted to 250,000 across the island. & Puerto Rico's housing secretary, Fernando Gil, says the number of \textbf{houses} destroyed by the hurricane totals about 70,000 so far, and homes with major damage have amounted to 250,000 across the island. \\ \hline
        EX2 FC\newline Typographic & Sabbir Khan. Sabbir's second movie, Heropanti starring Tiger Shroff \& Kriti Sanon\textbf{,} released on 23 may 2014. $\rightarrow$ Sabbir Khan directed a movie. & Sabbir Khan. Sabbir's second movie, Heropanti starring Tiger Shroff \& Kriti Sanon\textbf{?} released on 23 may 2014. $\rightarrow$ Sabbir Khan directed a movie. \\ \hline
        EX3 PR\newline Grammatical & Fastiggi and Goldstein have managed to make the problem even worse in their attempt to \textbf{explain} it away. & Fastiggi and Goldstein have managed to make the problem even worse in their attempt to \textbf{explained} it away. \\ \hline
        EX4 RD\newline Semantic-small & A few of the best cartoons \textbf{drawn \& shared in solidarity} with \#charliehebdo after yesterday's massacre \#jesuischarlie http://t.co/87et0xpnwr \newline @theinquisitr war profiteers x'd \#princessdiana \& dodifayed in \#paris. pushing \#france to join war on terror video $>>$ http://t.co/tysy8ys49w \newline @theinquisitr l'amérique se tient avec la france. \#jesuischarlie & A few of the best cartoons \textbf{contributed \& held in friendship} with \#charliehebdo after yesterday's massacre \#jesuischarlie http://t.co/87et0xpnwr \newline @theinquisitr war profiteers x'd \#princessdiana \& dodifayed in \#paris. pushing \#france to join war on terror video $>>$ http://t.co/tysy8ys49w \newline @theinquisitr l'amérique se tient avec la france. \#jesuischarlie \\ \hline
        EX5 FC \newline Semantic-large & Hannah and Her Sisters. Hannah and Her Sisters is a 1986 american comedy - drama film which tells the intertwined stories of an extended family over two years that begins and ends with a family thanksgiving dinner. $\rightarrow$ Hannah and Her Sisters is an American \textbf{1986} film. & Hannah and Her Sisters. Hannah and Her Sisters is a 1986 american comedy - drama film which tells the intertwined stories of an extended family over two years that begins and ends with a family thanksgiving dinner. $\rightarrow$ Hannah and Her Sisters is an American \textbf{1987} film. \\ \hline
        EX6 HN\newline Local & Aleppo completely back under government control (GPA) Aleppo – the Syrian Arab Army (SAA) has reported today that the entirety of \textbf{east} Aleppo is fully back under government control, meaning the city is now completely liberated. The SAA has completed the evacuations of anti-government fighters and civilians looking to flee with these groups as of today. This is a major victory for the Syrian forces in Aleppo coming after almost 4 years of fighting in the city. Thousands of people have already taken to the streets to celebrate the last of the terrorists inside the city leaving.\newline [347 words more]   & Aleppo completely back under government control (GPA) Aleppo – the Syrian Arab Army (SAA) has reported today that the entirety of \textbf{south} Aleppo is fully back under government control, meaning the city is now completely liberated. The SAA has completed the evacuations of anti-government fighters and civilians looking to flee with these groups as of today. This is a major victory for the Syrian forces in Aleppo coming after almost 4 years of fighting in the city. Thousands of people have already taken to the streets to celebrate the last of the terrorists inside the city leaving.\newline [347 words more] \\
    \hline
    \end{tabular}

    \label{tab:examples}
\end{table}

\section{Discussion}

\subsection{Reality check for credibility assessment}
\label{sec:realitycheck}
While one of the principles guiding the design of BODEGA has been a realistic simulation of the misinformation detection scenarios, this is possible only to an extent. Among the obstacles are low transparency of content management platforms \citep{Gorwa2020} and the vigorous growth of the methods of attack and defence in the NLP field. 

{Firstly, we have included only four victim models in our tests: BiLSTM, BERT and two Gemma variants, while in reality dozens of architectures for text classification are presented at every NLP conference, with a significant share specifically devoted to credibility assessment. However, the field has recently become surprisingly homogeneous, with the ambition to achieve the state of the art pushing researchers to reuse the common pretrained language models in virtually every application} \citep{Church2022}{. But these lookalike approaches share not only good performance, but also weaknesses. Thus we expect that, for example, the results of attacks on fine-tuned BERT will also apply to other solutions that use BERT as a representation layer. Moreover, the current architecture of BODEGA supports binary text classification models only. This means it can be extended to other similar tasks with a binary label output, e.g. sentiment analysis or detecting machine-generated text. But it cannot be used to assess robustness of models for machine translation or other language-generation tasks -- these would require a different approach. }

Secondly, we have re-used the attacks implemented in OpenAttack to have a comprehensive view of performance of different approaches. However, the field of AEs for NLP is relatively new, with the majority of publications emerging in the recent years, which makes it very likely that subsequent solutions will provide superior performance. With the creation of BODEGA as a universal evaluation framework, such comparisons become possible.

Thirdly, we need to consider the realism of evaluation measures. The AE evaluation framework assumes that if a modified text is very similar to the original, then the label (credible or not) still applies. Without this assumption, every evaluation would need to include manual re-annotation of the AEs. Fortunately, assessing semantic similarity  between two fragments of text is a necessary component of evaluation in many other NLP tasks, e.g. machine translation \citep{Lee2023}, and we can draw from that work. Apart from BLEURT, we have experimented with SBERT cross-encoders \citep{thakur-etal-2021-augmented} and unsupervised BERT Score \citep{DBLP:conf/iclr/ZhangKWWA20}, but haven't found decisive evidence for the superiority of any approach. However, the problem remains open. The investigation on how subtle changes in text can invert its meaning and subvert credibility assessment is particularly vivid in the fact-checking field \citep{Jaime2022}, but it is less explored for tasks involving multi-sentence inputs, e.g. news credibility. {An ideal measure of AE quality would take into account the characteristics of a text domain, assigning different impact to a given change depending on the nature of the text. This could be expressed by modifying the BODEGA score into a weighted score of the included factors and calibrating it by setting the weights for each text genre. However, to find the parameter values that accurately capture the human perception of acceptable changes, an annotation study would be necessary. We see this as a promising direction for future research.} {Moreover, the measures focusing on performance loss, e.g. computing the reduction in accuracy of the victim model under a specified modification might be worth investigating. However, an annotation study would be necessary as well, namely in order to establish the acceptable modification threshold for each task.}

Fourthly, we also assume that an attacker has a certain level of access to the victim classifier, being able to send unlimited queries and receive numerical scores reflecting its confidence, rather than a final decision. In practice, this is currently not the case, with platforms revealing almost nothing regarding their automatic content moderation processes. However, this may change in future due to regulatory pressure from the government organisations; cf., for example, the recently agreed EU \textit{Digital Services Act}\footnote{\url{https://ec.europa.eu/commission/presscorner/detail/en/qanda_20_2348}}.

Finally, we need to examine how realistic is that an attacker could freely modify any text included in our tasks. While this is trivial in the case of hyperpartisan news and propaganda recognition, where the entire input comes from a malicious actor, the other tasks require closer consideration. In case of rumour detection, the text includes, apart from the initial information, replies from other social media users. These can indeed be manipulated by sending replies from anonymous accounts and this scenario has been already explored in the AE literature \citep{Le2020}. In the case of fact checking, the text includes, apart from the verified claim, also the relevant snippets from the knowledge base. However, it can be modified as well, when (as is usually the case) the knowledge is based on Wikipedia, which is often a subject of malicious alterations, from vandalism \citep{Kiesel_Potthast_Hagen_Stein_2017} to the generation of entire hoax articles \citep{Kumar2016}.

To sum up, we argue that despite certain assumptions, the setup of a BODEGA framework is close enough to real-life conditions to give insights about the robustness of popular classifiers in this scenario. {BODEGA is already being used as a benchmark for new solutions that advance foundational AE generation methods tested here. Within the CheckThat! evaluation lab organised at CLEF 2024 }\citep{clef-checkthat:2024-lncs}{, focused on misinformation detection, Task 6 is devoted to measuring the robustness of credibility assessment. The evaluation of the AEs submitted by the task participants is based on the framework described here, with certain expansions } \citep{clef-checkthat:2024:task6}{\footnote{Note that the works cited here are in print at the time of writing.}.}

\subsection{Looking forward}

We see this study as a step towards the directions recognised in the ML literature beyond NLP. For example, in security-oriented applications, there is the need to bring the evaluation of AEs closer to realistic conditions \citep{Chen2022}. Some limitations, esp. number of queries to the model, make attacks much harder. Even beyond the security field, assessing robustness is crucial for ML models that are distributed as massively-used products. This exposes them to unexpected examples, even if not generated with explicit adversarial motive. Individual spectacular failures are expected to be disproportionately influential on public opinion of technology, including AI \citep{Mannes2020}, emphasising the importance of research on AEs.

Our work emphasises the need for taking into account the adversarial attacks when deploying text classifiers in adversarial scenarios, such as content filtering in social media. In many cases, changing just a few words in text can alter the decision of the models. We can recommend three ways to mitigate the associated risks. 

{Firstly, the vulnerability of ML models to adversarial examples indicates their output cannot be the only criterion in content-filtering systems. However, many AEs are quite transparent to humans, and and the manipulation could be easily noticed. This suggests that the sensitive scenarios could benefit from a cooperation between a human operator and a ML model. For example, a system that uses ML models for prioritising work of human operators instead of making final decision, is likely to be more robust than the ML model alone. Secondly, our work shows that the attack performance depends on the variety of factors, including dataset size, text length, victim architecture, etc. This makes it crucial to test every content filtering solution before its deployment using real-world data and state-of-the-art attackers. Thirdly, taking into account adversarial environment in the classifier design, e.g. through adversarial training, can limit the amount of adversarial examples it is vulnerable to.}

Finally, we need to acknowledge that the idea of using ML models for automatic moderation of user-generated content is not universally accepted, with some rejecting it as equivalent to censorship \citep{Llanso2020}, and calling for regulations in this area \citep{Meyer2019}. {Moreover, the recent changes in Twitter have served as an illustration on how relying on the automatic moderation to reduce operation costs }\citep{Paul2022}{ can result in more prevalent misinformation} \citep{Graham2023}.

\subsection{Using BODEGA}
{Beyond the exploration  of the current situation, we hope BODEGA will be useful for assessing the robustness of future classifiers and the effectiveness of new attacks. Towards this end, we make the software available openly\footnote{\url{https://github.com/piotrmp/BODEGA}}, allowing the replication of our experiments and evaluation of other solutions, both on the attack and the defence. Here, we also provide a handful of practical hints on how to use the software to perform such analysis in practice.

In order to \textbf{measure the robustness} of a classifier implemented in a particular scenario, the following is necessary:
}
\begin{enumerate}
    \item {Preparing a victim classifier. It can be based on the code in \texttt{runs/train\_victims.py}, which provides training of baseline classifiers  -- BiLSTM, BERT or GEMMA -- and only requires providing task-specific data. Otherwise, a completely different classifier can be included, as long as it implements the \texttt{OpenAttack.Classifier} interface. Note that both the classifier algorithm and training data will influence the robustness.}
    \item {Choosing an attacker. For this purpose, the results in table }\ref{tab:methodresults} {can be helpful, as they show the quality of the AEs as well as the number of queries. If the tested classifiers is deployed in a service that only allows a limited number of queries, this should be taken into account in simulating an attack.}
    \item {Evaluating an attack. This is performed by using the \texttt{runs/attack.py} script. Note that many of the attack methods consume significant computational resources and thus using a GPU device for both the victim and the attacker is recommended.}
    \item {Analysing the results. BODEGA will output both the overall evaluation results and all of the successful AEs, with the changes highlighted. It is recommended to analyse these manually, as the automatic meaning preservation methods have their limits, especially in specialised text domains.}
\end{enumerate}

{In order to \textbf{evaluate a new attack}, one needs to go through the following:}
\begin{enumerate}
    \item {Implement an attacker. It need to satisfy the \texttt{OpenAttack.attackers.ClassificationAttacker} interface, which sets out the procedure for finding AEs.} 
    \item {Choosing a victim. For the tasks and architectures tested here, the models are available for download from the BODEGA website. However, the \texttt{victims/transformer.py} script uses the \textit{HuggingFace} library, so a user can train a model with a newer architecture, as long as it is available through \texttt{AutoModelForSequenceClassification} interface.}
    \item {Evaluating an attack and analysing the results, as above.}
\end{enumerate}

{These are the most obvious usages of BODEGA, but other scenarios are possible as well, such as modifying the evaluation measure (BODEGA score) by improving the semantic similarity assessment, adding a different text classification task, linguistic inquiry into the generated AEs, cybersecurity-focused analyses, etc.}

\section{Conclusion}

Through this work, we have demonstrated that popular text classifiers, when applied for the purposes of misinformation detection, are vulnerable to manipulation through adversarial examples. We have discovered numerous cases where making a single barely perceptible change is enough to prevent a classifier from spotting non-credible information. Among the risk factors are large input lengths and the possibility of making numerous queries. Surprisingly, the classifiers trained on the basis of new state-of-the-art large language models are usually more vulnerable than their predecessors.

Nevertheless, the attack is never successful for every single instance and often entails changes that make text suspiciously malformed or ill-suited for the misinformation goal. This emphasises the need for thorough testing of the robustness of text classifiers at various stages of their development: from the initial design and experiments to the preparation for deployment, taking into account likely attack scenarios. We hope the BODEGA benchmark we contribute here, providing an environment for comprehensive and systematic tests, will be a useful tool in performing such analyses.


\section*{Acknowledgements}
This work is part of the ERINIA project, which has received funding from the European Union’s Horizon Europe research and innovation programme under grant agreement No 101060930.  Views and opinions expressed are however those of the author(s) only and do not necessarily reflect those of the European Union. Neither the European Union nor the granting authority can be held responsible for them. We also acknowledge the support from Departament de Recerca i Universitats de la Generalitat de Catalunya (ajuts SGR-Cat 2021) and the Spanish State Research Agency under the Maria de Maeztu Units of Excellence Programme (CEX2021-001195-M). {The computation for this study was made possible by the \textit{Google Cloud Platform} through research credits.}


\bibliographystyle{nlplike}
\bibliography{BODEGA}

\newpage
\section*{Appendix: full evaluation results}
This section contains full results obtained by evaluating the adversarial attacks on the misinformation classifiers. Specifically, we test the attack performance for:
\begin{itemize}
    \item four tasks (HN, PR, FC, RD),
    \item eight attackers (BAE, BERT-ATTACK, DeepWordBug, Genetic, SememePSO, PWWS, SCPN, TextFooler),
    \item four victims (BiLSTM, BERT, GEMMA2B, GEMMA7B),
    \item two scenarios (untargeted and targeted).
\end{itemize}

The following tables include the results for victims: BiLSTM (Table \ref{tab:BiLSTMresults}), BERT (Table \ref{tab:BERTresults}), GEMMA2B (Table \ref{tab:GEMMA2Bresults}) and GEMMA7B (Table \ref{tab:GEMMA7Bresults}).

\begin{table*}
    \small
    \begin{tabular}{rrrrrrrrrrrr}
    \hline
    &  & \multicolumn{5}{c}{\textbf{Untargeted}}  & \multicolumn{5}{c}{\textbf{Targeted}}\\
    \textbf{Task} & \textbf{Method} & \textbf{B.} & \textbf{Con} & \textbf{Sem} & \textbf{Char} & \textbf{Q.}  & \textbf{B.} & \textbf{con} & \textbf{sem} & \textbf{char} & \textbf{Q.} \\
    \hline
        HN & BAE  & {0.48} & {0.77} & {0.64} & {0.98} & {489.27} & {0.45} & {0.74} & {0.62} & {0.98} & {477.65} \\
         & BERT-ATTACK  & \textbf{0.64} & \textbf{0.98} & {0.66} & {0.99} & {487.85} & \textbf{0.61} & \textbf{0.96} & {0.65} & {0.99} & {565.05} \\
         & DeepWordBug  & {0.41} & {0.53} & \textbf{0.77} & \textbf{1.00} & {396.18} & {0.37} & {0.47} & \textbf{0.78} & \textbf{1.00} & {379.20} \\
         & Genetic  & {0.44} & {0.94} & {0.48} & {0.98} & {2029.31} & {0.42} & {0.90} & {0.47} & {0.98} & {2882.19} \\
         & SememePSO  & {0.21} & {0.42} & {0.50} & {0.99} & {313.51} & {0.14} & {0.28} & {0.49} & {0.99} & {361.38} \\
         & PWWS  & {0.44} & {0.93} & {0.48} & {0.99} & {2044.96} & {0.42} & {0.89} & {0.48} & {0.97} & {1994.95} \\
         & SCPN  & {0.00} & {0.94} & {0.08} & {0.02} & {11.86} & {0.00} & {0.95} & {0.08} & {0.02} & {11.83} \\
         & TextFooler  & {0.43} & {0.94} & {0.47} & {0.97} & {543.68} & {0.41} & {0.91} & {0.47} & {0.96} & {598.46} \\
    \hline
        PR & BAE  & {0.15} & {0.23} & {0.72} & {0.94} & {32.94} & {0.26} & {0.38} & {0.71} & {0.94} & {38.72} \\
         & BERT-ATTACK  & {0.53} & {0.80} & {0.72} & {0.91} & {61.41} & \textbf{0.66} & {0.94} & {0.74} & {0.94} & {50.14} \\
         & DeepWordBug  & {0.29} & {0.38} & \textbf{0.79} & \textbf{0.96} & {27.45} & {0.56} & {0.72} & \textbf{0.81} & \textbf{0.96} & {35.30} \\
         & Genetic  & \textbf{0.54} & \textbf{0.88} & {0.67} & {0.89} & {782.15} & {0.62} & {0.94} & {0.71} & {0.93} & {802.20} \\
         & SememePSO  & {0.47} & {0.76} & {0.68} & {0.89} & {85.34} & {0.60} & {0.92} & {0.71} & {0.92} & {69.62} \\
         & PWWS  & {0.53} & {0.84} & {0.69} & {0.90} & {130.85} & {0.63} & {0.92} & {0.73} & {0.94} & {168.60} \\
         & SCPN  & {0.12} & {0.55} & {0.39} & {0.50} & {11.55} & {0.20} & \textbf{0.98} & {0.37} & {0.48} & {11.98} \\
         & TextFooler  & {0.51} & {0.85} & {0.67} & {0.88} & {52.59} & {0.63} & {0.94} & {0.72} & {0.92} & {54.62} \\
    \hline
        FC & BAE  & {0.36} & {0.55} & {0.69} & {0.96} & {77.76} & {0.32} & {0.48} & {0.69} & {0.96} & {73.43} \\
         & BERT-ATTACK  & {0.60} & {0.86} & {0.73} & {0.95} & {132.80} & \textbf{0.59} & {0.85} & {0.73} & {0.96} & {123.24} \\
         & DeepWordBug  & {0.48} & {0.58} & \textbf{0.85} & \textbf{0.98} & {54.36} & {0.54} & {0.64} & \textbf{0.85} & \textbf{0.98} & {50.72} \\
         & Genetic  & \textbf{0.61} & \textbf{0.90} & {0.71} & {0.95} & {840.99} & {0.57} & {0.88} & {0.69} & {0.94} & {1015.44} \\
         & SememePSO  & {0.53} & {0.76} & {0.72} & {0.96} & {112.84} & {0.46} & {0.67} & {0.72} & {0.96} & {132.28} \\
         & PWWS  & {0.57} & {0.82} & {0.73} & {0.96} & {221.60} & {0.50} & {0.73} & {0.71} & {0.95} & {211.05} \\
         & SCPN  & {0.08} & {0.75} & {0.29} & {0.32} & {11.75} & {0.11} & \textbf{1.00} & {0.30} & {0.35} & {12.00} \\
         & TextFooler  & {0.55} & {0.82} & {0.71} & {0.94} & {98.31} & {0.50} & {0.75} & {0.70} & {0.94} & {99.98} \\
    \hline
        RD & BAE  & {0.09} & {0.21} & {0.43} & {0.98} & {312.77} & {0.27} & {0.64} & {0.43} & {0.98} & {123.16} \\
         & BERT-ATTACK  & {0.29} & \textbf{0.79} & {0.41} & {0.89} & {985.52} & {0.43} & {0.95} & {0.46} & {0.97} & {130.64} \\
         & DeepWordBug  & {0.16} & {0.24} & \textbf{0.68} & \textbf{0.99} & {232.75} & \textbf{0.62} & {0.91} & \textbf{0.69} & \textbf{0.99} & {153.61} \\
         & Genetic  & \textbf{0.32} & {0.71} & {0.47} & {0.96} & {3150.24} & {0.44} & \textbf{0.96} & {0.48} & {0.95} & {1355.52} \\
         & SememePSO  & {0.15} & {0.31} & {0.48} & {0.97} & {314.63} & {0.32} & {0.67} & {0.50} & {0.97} & {185.47} \\
         & PWWS  & {0.29} & {0.64} & {0.47} & {0.97} & {1059.07} & {0.44} & {0.95} & {0.48} & {0.95} & {742.12} \\
         & SCPN  & {0.01} & {0.55} & {0.17} & {0.09} & {11.53} & {0.02} & {0.84} & {0.15} & {0.12} & {11.84} \\
         & TextFooler  & {0.24} & {0.64} & {0.41} & {0.87} & {639.97} & {0.44} & \textbf{0.96} & {0.48} & {0.96} & {184.97} \\
    \hline
    \end{tabular}
    \caption{The results of adversarial attacks on the \textbf{BiLSTM} classifier in four misinformation detection tasks in untargeted and targeted scenario. Evaluation measures include BODEGA score (B.), confusion score (con), semantic score (sem), character score (char) and number of queries to the attacked model (Q.). The best score in each task and scenario is in boldface. }
    \label{tab:BiLSTMresults}
\end{table*}

\begin{table*}
    \small
    \begin{tabular}{rrrrrrrrrrrr}
    \hline
    &  & \multicolumn{5}{c}{\textbf{Untargeted}}  & \multicolumn{5}{c}{\textbf{Targeted}}\\
    \textbf{Task} & \textbf{Method} & \textbf{B.} & \textbf{Con} & \textbf{Sem} & \textbf{Char} & \textbf{Q.}  & \textbf{B.} & \textbf{con} & \textbf{sem} & \textbf{char} & \textbf{Q.} \\
    \hline
        HN & BAE  & {0.34} & {0.60} & {0.58} & {0.96} & {606.83} & {0.18} & {0.34} & {0.57} & {0.95} & {713.42} \\
         & BERT-ATTACK  & \textbf{0.60} & \textbf{0.96} & {0.64} & {0.97} & {648.41} & \textbf{0.57} & \textbf{0.95} & {0.62} & {0.96} & {753.91} \\
         & DeepWordBug  & {0.22} & {0.29} & \textbf{0.78} & \textbf{1.00} & {395.94} & {0.15} & {0.20} & \textbf{0.78} & \textbf{1.00} & {389.81} \\
         & Genetic  & {0.40} & {0.86} & {0.47} & {0.98} & {2713.80} & {0.30} & {0.71} & {0.44} & {0.97} & {4502.51} \\
         & SememePSO  & {0.16} & {0.34} & {0.50} & {0.99} & {341.70} & {0.05} & {0.12} & {0.44} & {0.99} & {417.99} \\
         & PWWS  & {0.38} & {0.82} & {0.47} & {0.98} & {2070.78} & {0.27} & {0.64} & {0.44} & {0.95} & {2107.02} \\
         & SCPN  & {0.00} & {0.92} & {0.08} & {0.02} & {11.84} & {0.00} & \textbf{0.95} & {0.09} & {0.02} & {11.89} \\
         & TextFooler  & {0.39} & {0.92} & {0.44} & {0.94} & {660.52} & {0.32} & {0.85} & {0.41} & {0.90} & {850.79} \\
    \hline
        PR & BAE  & {0.11} & {0.18} & {0.69} & {0.94} & {33.96} & {0.13} & {0.20} & {0.68} & {0.94} & {45.68} \\
         & BERT-ATTACK  & {0.43} & {0.70} & {0.68} & {0.90} & {80.16} & \textbf{0.50} & {0.79} & {0.69} & {0.92} & {99.95} \\
         & DeepWordBug  & {0.28} & {0.36} & \textbf{0.79} & \textbf{0.96} & {27.43} & {0.50} & {0.64} & \textbf{0.81} & \textbf{0.96} & {36.04} \\
         & Genetic  & \textbf{0.50} & \textbf{0.84} & {0.65} & {0.89} & {962.40} & {0.49} & \textbf{0.84} & {0.65} & {0.89} & {1211.56} \\
         & SememePSO  & {0.41} & {0.68} & {0.66} & {0.90} & {96.17} & {0.35} & {0.53} & {0.71} & {0.91} & {173.71} \\
         & PWWS  & {0.47} & {0.75} & {0.68} & {0.91} & {131.92} & {0.44} & {0.72} & {0.68} & {0.89} & {179.68} \\
         & SCPN  & {0.09} & {0.47} & {0.36} & {0.46} & {11.47} & {0.11} & {0.79} & {0.32} & {0.39} & {11.79} \\
         & TextFooler  & {0.43} & {0.77} & {0.64} & {0.87} & {57.94} & {0.46} & {0.77} & {0.66} & {0.89} & {77.81} \\
    \hline
        FC & BAE  & {0.34} & {0.51} & {0.70} & {0.96} & {80.69} & {0.18} & {0.27} & {0.70} & {0.94} & {92.47} \\
         & BERT-ATTACK  & \textbf{0.53} & {0.77} & {0.73} & {0.95} & {146.73} & \textbf{0.41} & {0.62} & {0.71} & {0.93} & {207.23} \\
         & DeepWordBug  & {0.44} & {0.53} & \textbf{0.84} & \textbf{0.98} & {54.32} & {0.22} & {0.27} & \textbf{0.85} & \textbf{0.98} & {52.31} \\
         & Genetic  & {0.52} & {0.79} & {0.70} & {0.95} & {1215.19} & {0.39} & {0.63} & {0.66} & {0.92} & {1808.08} \\
         & SememePSO  & {0.44} & {0.64} & {0.71} & {0.96} & {148.20} & {0.25} & {0.37} & {0.70} & {0.94} & {230.58} \\
         & PWWS  & {0.48} & {0.69} & {0.72} & {0.96} & {225.27} & {0.31} & {0.47} & {0.70} & {0.94} & {226.78} \\
         & SCPN  & {0.09} & \textbf{0.90} & {0.29} & {0.31} & {11.90} & {0.09} & \textbf{0.97} & {0.29} & {0.30} & {11.97} \\
         & TextFooler  & {0.46} & {0.70} & {0.70} & {0.93} & {106.13} & {0.29} & {0.49} & {0.65} & {0.88} & {131.88} \\
    \hline
        RD & BAE  & {0.07} & {0.18} & {0.41} & {0.98} & {313.01} & {0.18} & {0.44} & {0.42} & {0.98} & {196.69} \\
         & BERT-ATTACK  & {0.18} & {0.44} & {0.43} & {0.96} & {774.31} & {0.30} & {0.69} & {0.45} & {0.97} & {366.14} \\
         & DeepWordBug  & {0.16} & {0.23} & \textbf{0.70} & \textbf{0.99} & {232.74} & \textbf{0.39} & {0.56} & \textbf{0.70} & \textbf{0.99} & {174.03} \\
         & Genetic  & \textbf{0.20} & \textbf{0.46} & {0.45} & {0.96} & {4425.11} & {0.35} & {0.79} & {0.46} & {0.95} & {2266.91} \\
         & SememePSO  & {0.10} & {0.21} & {0.46} & {0.97} & {345.89} & {0.27} & {0.57} & {0.49} & {0.96} & {233.88} \\
         & PWWS  & {0.16} & {0.38} & {0.45} & {0.95} & {1105.99} & {0.32} & {0.75} & {0.45} & {0.93} & {838.83} \\
         & SCPN  & {0.01} & {0.38} & {0.16} & {0.10} & {11.35} & {0.02} & \textbf{0.90} & {0.15} & {0.10} & {11.90} \\
         & TextFooler  & {0.16} & {0.41} & {0.43} & {0.91} & {657.15} & {0.31} & {0.70} & {0.47} & {0.96} & {358.37} \\
    \hline
    \end{tabular}
    \caption{The results of adversarial attacks on the \textbf{BERT} classifier (see the caption of the previous table). }
    \label{tab:BERTresults}
\end{table*}

\begin{table*}
    \small
    \begin{tabular}{rrrrrrrrrrrr}
    \hline
    &  & \multicolumn{5}{c}{\textbf{Untargeted}}  & \multicolumn{5}{c}{\textbf{Targeted}}\\
    \textbf{Task} & \textbf{Method} & \textbf{B.} & \textbf{Con} & \textbf{Sem} & \textbf{Char} & \textbf{Q.}  & \textbf{B.} & \textbf{con} & \textbf{sem} & \textbf{char} & \textbf{Q.} \\
    \hline
        HN & BAE  & {0.36} & {0.61} & {0.60} & {0.97} & {583.38} & {0.38} & {0.66} & {0.59} & {0.97} & {545.68} \\
         & BERT-ATTACK  & \textbf{0.55} & \textbf{0.91} & {0.62} & {0.96} & {942.98} & \textbf{0.57} & {0.95} & {0.62} & {0.97} & {761.53} \\
         & DeepWordBug  & {0.24} & {0.31} & \textbf{0.78} & \textbf{1.00} & {385.91} & {0.27} & {0.34} & \textbf{0.78} & \textbf{1.00} & {380.58} \\
         & Genetic  & {0.40} & {0.84} & {0.48} & {0.99} & {1176.75} & {0.44} & {0.93} & {0.48} & {0.99} & {894.47} \\
         & SememePSO  & {0.26} & {0.54} & {0.48} & {1.00} & {236.96} & {0.32} & {0.66} & {0.49} & {0.99} & {185.02} \\
         & PWWS  & {0.38} & {0.82} & {0.48} & {0.98} & {2021.92} & {0.45} & {0.97} & {0.47} & {0.98} & {1980.32} \\
         & SCPN  & {0.00} & {0.69} & {0.09} & {0.02} & {11.62} & {0.00} & \textbf{0.99} & {0.10} & {0.02} & {11.88} \\
         & TextFooler  & {0.32} & {0.70} & {0.47} & {0.97} & {834.56} & {0.41} & {0.89} & {0.47} & {0.98} & {583.29} \\
    \hline
        PR & BAE  & {0.16} & {0.23} & {0.72} & {0.94} & {32.64} & {0.21} & {0.32} & {0.69} & {0.94} & {43.68} \\
         & BERT-ATTACK  & {0.46} & {0.72} & {0.70} & {0.91} & {77.50} & \textbf{0.55} & {0.85} & {0.70} & {0.92} & {94.05} \\
         & DeepWordBug  & {0.14} & {0.19} & \textbf{0.79} & \textbf{0.96} & {27.26} & {0.29} & {0.38} & \textbf{0.80} & \textbf{0.97} & {36.04} \\
         & Genetic  & \textbf{0.50} & \textbf{0.84} & {0.66} & {0.90} & {876.06} & {0.51} & \textbf{0.86} & {0.65} & {0.91} & {1077.41} \\
         & SememePSO  & {0.42} & {0.69} & {0.67} & {0.90} & {93.41} & {0.35} & {0.57} & {0.67} & {0.91} & {180.15} \\
         & PWWS  & {0.45} & {0.74} & {0.67} & {0.91} & {132.13} & {0.44} & {0.73} & {0.66} & {0.90} & {179.97} \\
         & SCPN  & {0.14} & {0.62} & {0.40} & {0.51} & {11.62} & {0.15} & {0.74} & {0.36} & {0.45} & {11.74} \\
         & TextFooler  & {0.36} & {0.63} & {0.65} & {0.87} & {68.64} & {0.37} & {0.66} & {0.63} & {0.87} & {97.49} \\
    \hline
        FC & BAE  & {0.34} & {0.51} & {0.70} & {0.96} & {79.43} & {0.17} & {0.26} & {0.69} & {0.96} & {92.43} \\
         & BERT-ATTACK  & \textbf{0.57} & \textbf{0.83} & {0.72} & {0.94} & {192.25} & \textbf{0.50} & {0.76} & {0.70} & {0.92} & {254.22} \\
         & DeepWordBug  & {0.07} & {0.09} & \textbf{0.83} & \textbf{0.98} & {53.88} & {0.06} & {0.08} & \textbf{0.84} & \textbf{0.98} & {52.02} \\
         & Genetic  & {0.46} & {0.68} & {0.71} & {0.96} & {486.65} & {0.31} & {0.50} & {0.68} & {0.93} & {626.83} \\
         & SememePSO  & {0.42} & {0.62} & {0.71} & {0.96} & {155.18} & {0.22} & {0.33} & {0.70} & {0.94} & {242.62} \\
         & PWWS  & {0.43} & {0.63} & {0.72} & {0.96} & {228.43} & {0.22} & {0.33} & {0.71} & {0.95} & {230.18} \\
         & SCPN  & {0.06} & {0.51} & {0.31} & {0.34} & {11.51} & {0.11} & \textbf{1.00} & {0.31} & {0.34} & {12.00} \\
         & TextFooler  & {0.43} & {0.65} & {0.71} & {0.94} & {120.38} & {0.24} & {0.38} & {0.67} & {0.92} & {137.26} \\
    \hline
        RD & BAE  & {0.13} & {0.31} & {0.42} & {0.98} & {298.30} & {0.19} & {0.47} & {0.41} & {0.97} & {170.87} \\
         & BERT-ATTACK  & \textbf{0.30} & \textbf{0.73} & {0.43} & {0.95} & {703.07} & {0.41} & {0.93} & {0.45} & {0.96} & {259.36} \\
         & DeepWordBug  & {0.10} & {0.15} & \textbf{0.69} & \textbf{0.99} & {238.97} & {0.24} & {0.35} & \textbf{0.69} & \textbf{0.99} & {161.87} \\
         & Genetic  & {0.24} & {0.54} & {0.46} & {0.96} & {1647.75} & \textbf{0.41} & {0.90} & {0.47} & {0.96} & {1108.21} \\
         & SememePSO  & {0.12} & {0.27} & {0.46} & {0.98} & {323.73} & {0.24} & {0.53} & {0.47} & {0.97} & {214.95} \\
         & PWWS  & {0.21} & {0.46} & {0.47} & {0.97} & {1124.29} & {0.40} & {0.89} & {0.47} & {0.96} & {774.03} \\
         & SCPN  & {0.01} & {0.38} & {0.16} & {0.11} & {11.38} & {0.02} & \textbf{0.95} & {0.16} & {0.13} & {11.95} \\
         & TextFooler  & {0.19} & {0.44} & {0.45} & {0.94} & {640.54} & {0.38} & {0.86} & {0.47} & {0.94} & {260.58} \\
    \hline    \end{tabular}
    \caption{The results of adversarial attacks on the \textbf{GEMMA2B} classifier (see the caption of the previous table).}
    \label{tab:GEMMA2Bresults}
\end{table*}

\begin{table*}
    \small
    \begin{tabular}{rrrrrrrrrrrr}
    \hline
    &  & \multicolumn{5}{c}{\textbf{Untargeted}}  & \multicolumn{5}{c}{\textbf{Targeted}}\\
    \textbf{Task} & \textbf{Method} & \textbf{B.} & \textbf{Con} & \textbf{Sem} & \textbf{Char} & \textbf{Q.}  & \textbf{B.} & \textbf{con} & \textbf{sem} & \textbf{char} & \textbf{Q.} \\
    \hline
        HN & BAE  & {0.25} & {0.42} & {0.62} & {0.97} & {678.07} & {0.11} & {0.18} & {0.62} & {0.98} & {740.02} \\
         & BERT-ATTACK  & \textbf{0.45} & \textbf{0.75} & {0.62} & {0.95} & {1560.76} & \textbf{0.31} & {0.55} & {0.59} & {0.93} & {2313.33} \\
         & DeepWordBug  & {0.14} & {0.18} & \textbf{0.79} & \textbf{1.00} & {385.78} & {0.04} & {0.05} & \textbf{0.81} & \textbf{1.00} & {377.38} \\
         & Genetic  & {0.30} & {0.61} & {0.49} & {0.99} & {1042.31} & {0.19} & {0.42} & {0.47} & {0.98} & {1159.68} \\
         & SememePSO  & {0.15} & {0.29} & {0.51} & {0.99} & {344.23} & {0.07} & {0.14} & {0.50} & {1.00} & {403.99} \\
         & PWWS  & {0.28} & {0.60} & {0.48} & {0.98} & {2068.81} & {0.17} & {0.40} & {0.45} & {0.96} & {2095.51} \\
         & SCPN  & {0.00} & {0.73} & {0.09} & {0.02} & {11.66} & {0.00} & \textbf{0.65} & {0.09} & {0.03} & {11.54} \\
         & TextFooler  & {0.22} & {0.51} & {0.46} & {0.94} & {1132.87} & {0.16} & {0.40} & {0.43} & {0.91} & {1284.86} \\
    \hline
        PR & BAE  & {0.13} & {0.19} & {0.72} & {0.94} & {33.69} & {0.22} & {0.32} & {0.72} & {0.94} & {44.26} \\
         & BERT-ATTACK  & {0.42} & {0.67} & {0.69} & {0.91} & {86.54} & \textbf{0.53} & {0.78} & {0.72} & {0.93} & {110.32} \\
         & DeepWordBug  & {0.09} & {0.12} & \textbf{0.79} & \textbf{0.96} & {27.19} & {0.23} & {0.29} & \textbf{0.81} & \textbf{0.97} & {35.37} \\
         & Genetic  & \textbf{0.44} & \textbf{0.78} & {0.63} & {0.88} & {925.58} & {0.49} & {0.85} & {0.64} & {0.89} & {845.27} \\
         & SememePSO  & {0.34} & {0.58} & {0.65} & {0.88} & {123.13} & {0.38} & {0.62} & {0.68} & {0.91} & {165.24} \\
         & PWWS  & {0.38} & {0.64} & {0.65} & {0.89} & {133.92} & {0.43} & {0.71} & {0.66} & {0.90} & {179.77} \\
         & SCPN  & {0.10} & {0.51} & {0.36} & {0.47} & {11.52} & {0.16} & \textbf{0.86} & {0.36} & {0.46} & {11.86} \\
         & TextFooler  & {0.34} & {0.62} & {0.62} & {0.86} & {69.88} & {0.40} & {0.65} & {0.68} & {0.90} & {86.74} \\
    \hline
        FC & BAE  & {0.36} & {0.55} & {0.68} & {0.96} & {76.44} & {0.20} & {0.31} & {0.68} & {0.95} & {87.62} \\
         & BERT-ATTACK  & \textbf{0.58} & \textbf{0.85} & {0.72} & {0.95} & {141.70} & \textbf{0.52} & {0.80} & {0.69} & {0.93} & {173.86} \\
         & DeepWordBug  & {0.04} & {0.06} & \textbf{0.81} & \textbf{0.98} & {53.84} & {0.02} & {0.03} & \textbf{0.81} & \textbf{0.99} & {51.55} \\
         & Genetic  & {0.47} & {0.73} & {0.68} & {0.94} & {842.16} & {0.34} & {0.58} & {0.64} & {0.91} & {1205.46} \\
         & SememePSO  & {0.39} & {0.58} & {0.70} & {0.96} & {164.01} & {0.23} & {0.35} & {0.69} & {0.95} & {228.84} \\
         & PWWS  & {0.42} & {0.62} & {0.70} & {0.96} & {228.62} & {0.24} & {0.37} & {0.68} & {0.94} & {227.46} \\
         & SCPN  & {0.06} & {0.50} & {0.31} & {0.34} & {11.50} & {0.11} & \textbf{1.00} & {0.31} & {0.34} & {12.00} \\
         & TextFooler  & {0.41} & {0.63} & {0.69} & {0.94} & {114.26} & {0.24} & {0.40} & {0.66} & {0.91} & {132.95} \\
    \hline
        RD & BAE  & {0.10} & {0.26} & {0.41} & {0.97} & {318.74} & {0.21} & {0.53} & {0.40} & {0.97} & {168.33} \\
         & BERT-ATTACK  & \textbf{0.21} & \textbf{0.52} & {0.43} & {0.95} & {977.27} & \textbf{0.44} & \textbf{1.00} & {0.46} & {0.96} & {202.18} \\
         & DeepWordBug  & {0.08} & {0.12} & \textbf{0.70} & \textbf{0.99} & {238.93} & {0.21} & {0.31} & \textbf{0.70} & \textbf{0.99} & {156.29} \\
         & Genetic  & {0.18} & {0.40} & {0.47} & {0.97} & {1197.40} & {0.42} & {0.93} & {0.46} & {0.96} & {1531.17} \\
         & SememePSO  & {0.11} & {0.23} & {0.48} & {0.98} & {336.53} & {0.26} & {0.58} & {0.46} & {0.97} & {208.59} \\
         & PWWS  & {0.17} & {0.37} & {0.47} & {0.97} & {1139.02} & {0.39} & {0.88} & {0.46} & {0.96} & {762.62} \\
         & SCPN  & {0.01} & {0.34} & {0.17} & {0.13} & {11.34} & {0.03} & {0.98} & {0.17} & {0.14} & {11.98} \\
         & TextFooler  & {0.15} & {0.36} & {0.45} & {0.95} & {679.12} & {0.40} & {0.90} & {0.46} & {0.95} & {249.94} \\
    \hline 
    \end{tabular}
    \caption{The results of adversarial attacks on the \textbf{GEMMA7B classifier} (see the caption of the previous table).}
    \label{tab:GEMMA7Bresults}
\end{table*}

\label{lastpage}

\end{document}